%% file: main.tex
\documentclass[10pt,twocolumn,letterpaper]{article}

\usepackage[pagenumbers]{cvpr}      % To produce the REVIEW version
\input{preamble}

\definecolor{cvprblue}{rgb}{0.21,0.49,0.74}
\usepackage[pagebackref,breaklinks,colorlinks,allcolors=cvprblue]{hyperref}

\def\confName{CVPR}
\def\confYear{2027}

\title{JointGV: Structure-Aware Correction of Visual World Models with Graph Dynamics for Long-Horizon Modeling}

\author{
Tian Lyu\textsuperscript{1,\textdagger} \quad
Yichen Jiang\textsuperscript{2,\textdagger} \quad
Ziyuan Cheng\textsuperscript{3} \quad
Yunda Yan\textsuperscript{4} \quad
Xiangyu Pan\textsuperscript{2}\\
Junjie Jiang\textsuperscript{2} \quad
Jialin Yu\textsuperscript{5,1} \quad
Boyu Ning\textsuperscript{6} \quad
Pietro Li\`o\textsuperscript{7} \quad
Fan Mo\textsuperscript{2,*}\\[4pt]
{\small
\textsuperscript{1}University of Oxford \quad
\textsuperscript{2}University of Electronic Science and Technology of China}\\
{\small
\textsuperscript{3}Monash University \quad
\textsuperscript{4}University College London, University of London}\\
{\small
\textsuperscript{5}University of Bath \quad
\textsuperscript{6}KTH Royal Institute of Technology \quad
\textsuperscript{7}University of Cambridge}\\[2pt]
{\small
\textsuperscript{\textdagger}These authors contributed equally to this work.}\\
{\small
\textsuperscript{*}Corresponding author:
\texttt{fan.mo@uestc.edu.cn}}
}

\begin{document}
\maketitle
\input{sec/0_abstract}

\begin{center}
\footnotesize
\textbf{Code: }
\url{https://github.com/JasonLvernex/Joint_Graph_Vedio_Latent_Dynamic_World_Model}
\end{center}
\vspace{-4pt}

\input{sec/intro}

\input{sec/Method}

\input{sec/Experiment_Result}
{
    \small
    \bibliographystyle{ieeenat_fullname}
    \bibliography{main}
}

\input{sec/X_suppl}

\end{document}

%% file: preamble.tex
\usepackage{placeins}
\usepackage{needspace}
\usepackage{float}

%% file: sec/0_abstract.tex
\begin{abstract}

Visual world models predict future visual states conditioned on actions, but recursive errors in entity poses and spatial relations limit their long-horizon prediction for robot planning and control. We ask whether a separately predicted relational state can correct this drift without restructuring the visual model. This is challenging because graph nodes and visual states share neither elements nor a feature space. We propose Joint Graph--Visual Dynamics Correction (JointGV), a plug-in based on a cross-granularity dynamics hypothesis: despite their different representations, graph-node and visual-state changes describe the same physical transition. JointGV rolls out a task-conditioned scene graph alongside the visual world model, couples their predicted changes, and combines the resulting graph-motion context with graph-predicted node positions to compute a bounded residual. Graph-learning and correction gradients do not propagate into the original visual dynamics. We attach JointGV to DINO-WM (spatial tokens) and DreamerV3 (a global latent state), evaluating both on DMC Cheetah Run and Walker Walk across three seeds and extending the evaluation to contact-rich manipulation with DMC Finger Turn Hard. Across 32-step rollouts, JointGV improves structural and visual-prediction accuracy, reducing joint-position error by 11--28\%. For spatial tokens, learned graph--visual change coupling gives lower error than broadcasting the same mean graph-motion context to every token, particularly when non-adjacent parts appear close together in the image. On real-robot inverted-pendulum sequences, JointGV reduces DINO-WM's 32-step pole-angle error from $12.56^\circ$ to $6.69^\circ$ and improves DreamerV3 and LeWorldModel as well. In simulated inverted-pendulum control, an actor trained with JointGV-corrected rollouts completes 46 of 50 episodes, compared with none using the original rollouts.
% JointGV combines graph--visual change alignment with graph-predicted node positions to correct long-horizon visual rollouts, providing a design that may extend to other visual world models while preserving their original visual dynamics.
\end{abstract}

%% file: sec/intro.tex
\section{Introduction}
\label{sec:introduction}

\begin{figure}[!t]
    \centering
    \includegraphics[width=0.9\linewidth]{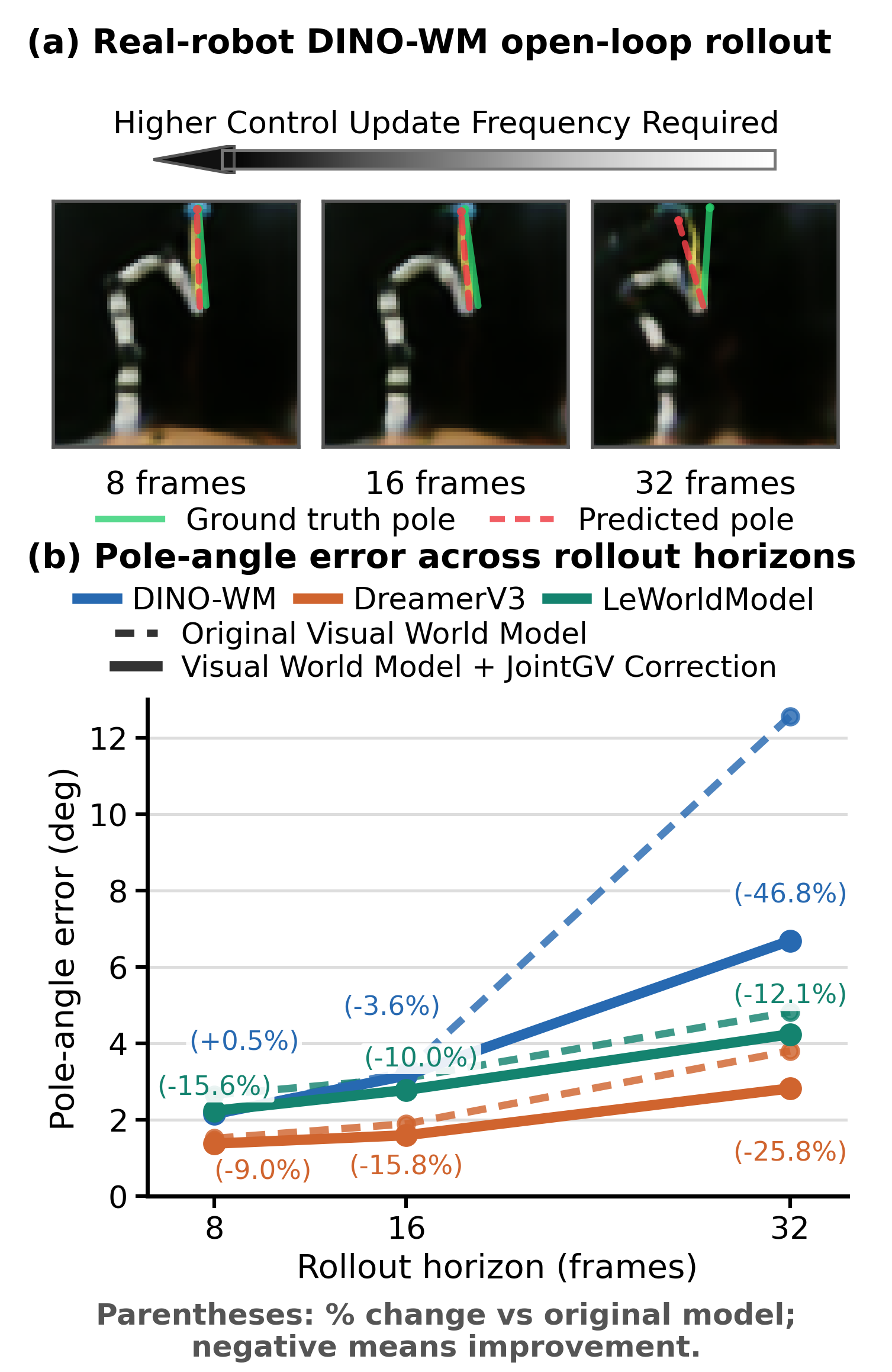}
    \caption{\textbf{Motivation.} (a) DINO-WM open-loop predictions initialized from real-robot inverted-pendulum observations increasingly deviate from the observed pole. (b) Pole-angle error rises with rollout horizon for all three models, while JointGV reduces this structural drift.}
    \label{fig:motivation}
     \vspace{-21pt}
\end{figure}

Visual world models predict how a robot and its environment change under an action. Robot planning evaluates multi-step action sequences, making recursive rollout necessary~\cite{wu2023daydreamer,hafner2025mastering,zhou2024dinowm}. Reusing each predicted state as the next input, however, accumulates errors in entity poses and spatial relations. Figure~\ref{fig:motivation} shows this structural drift in open-loop predictions initialized from RGB observations of a physical inverted-pendulum robot: the predicted pole progressively deviates and blurs, while its angle error increases with rollout horizon across three visual world models.

Existing approaches reduce long-horizon error by changing the native predictive representation~\cite{xia2026cloning,zhang2026geoworld}, using auxiliary trajectory supervision~\cite{yang2026fourDwam}, learning residuals within the original latent pathway~\cite{lanier2026redraw,soni2026sandwich}, or replacing the visual state with object- or graph-centric dynamics~\cite{lin2020gswm,feng2023daft}. These approaches modify the visual representation, derive correction from the same latent pathway, or require a separately trained structured world model. They do not address how an independently predicted relational rollout can correct an existing visual state when their elements and feature spaces do not correspond.

Our key hypothesis is that, although graph nodes and visual states differ in granularity and content, their action-induced changes describe the same physical transition. We propose \emph{Joint Graph--Visual Dynamics Correction} (JointGV), which rolls out a task-conditioned scene graph alongside an existing visual world model and couples their predicted changes in a shared space. Graph-predicted node positions provide an additional spatial reference, and a bounded, gated residual corrects the visual prediction. Graph-learning and correction gradients do not propagate into the original visual encoder or dynamics.

Our contributions are: (1) a cross-granularity graph--visual correction interface for predictive states without shared elements or feature spaces; (2) a plug-in architecture that uses a separately predicted relational rollout while preserving the original visual dynamics; and (3) three-seed evaluation across four DMC locomotion pairings and a contact-rich DMC manipulation task, real-robot prediction, simulated control, and analyses identifying when learned graph--visual coupling is most useful.

\section{Related Work}
\label{sec:related_work}

\paragraph{Visual world models for robot learning.}
Visual world models learn action-conditioned states for planning and control. DayDreamer and DreamerV3 learn behaviors in latent imagination~\cite{wu2023daydreamer,hafner2025mastering}. DINO-WM predicts DINOv2 patch features, IRASim predicts robot--object videos, and LeWorldModel predicts image embeddings~\cite{oquab2023dinov2,zhou2024dinowm,zhu2025irasim,maes2026leworldmodel}. 

\paragraph{Object-centric and graph-structured world models.}
Graph networks model physical systems through entity states and relational message passing~\cite{sanchez2018graph,sanchez2020learning,liu2026graphworldmodels}. C-SWM, G-SWM, SlotFormer, and DAFT-RL build object interactions into the predictive state~\cite{kipf2020contrastive,lin2020gswm,wu2023slotformer,feng2023daft}, while recent methods extend structured prediction to multimodal, recurrent, and evolving graphs~\cite{feng2025graphworld,wang2025dyno,berkes2025graphdreamer,schutz2026gdm}. These methods use the structured state as part of the world model itself. JointGV instead preserves an existing visual state and uses an independently predicted graph rollout as a correction source.

\paragraph{Structure and residual correction of world models.}
GRWM and GeoWorld alter the geometry of the native predictive representation~\cite{xia2026cloning,zhang2026geoworld}. 4D-WAM aligns spatially corresponding video and trajectory-field changes as auxiliary training supervision~\cite{yang2026fourDwam}. ReDRAW and Sandwich-Residuals instead learn residuals within the original latent pathway to adapt to dynamics or distribution shift~\cite{lanier2026redraw,soni2026sandwich}. Structure-guided video prediction generates frames from predicted poses or keypoints~\cite{villegas2017hierarchical,minderer2019unsupervised}, and PhyDNet separates physical dynamics from residual factors~\cite{leguen2020disentangling}; these methods train the visual predictor around the structured state. JointGV addresses a different interface: it couples an independently predicted scene-graph rollout with visual states that have no shared indexing or feature space, then corrects the visual prediction without propagating correction gradients into the original visual model.

%% file: sec/Method.tex
\section{Method}
\label{sec:method}

\subsection{Overview}
\label{sec:method_overview}

Let $\mathbf{Z}_t$ denote a spatial-token or global-latent visual state, and let $\overline{\mathbf{Z}}_{t+1}$ be its next-state prediction from the original visual world model. JointGV predicts graph-node features $\widehat{\mathbf{N}}_{t+1}$ in parallel. Their one-step changes are
\begin{equation}
\Delta\overline{\mathbf{Z}}_t
=
\overline{\mathbf{Z}}_{t+1}-\mathbf{Z}_t,
\qquad
\Delta\widehat{\mathbf{N}}_t
=
\widehat{\mathbf{N}}_{t+1}-\mathbf{N}_t,
\label{eq:state_increments}
\end{equation}
where the bar denotes the original visual prediction, the hat denotes the graph-dynamics prediction, and $\Delta$ denotes a one-step change.

\begin{figure*}[t]
    \centering
    \includegraphics[width=0.8\textwidth]{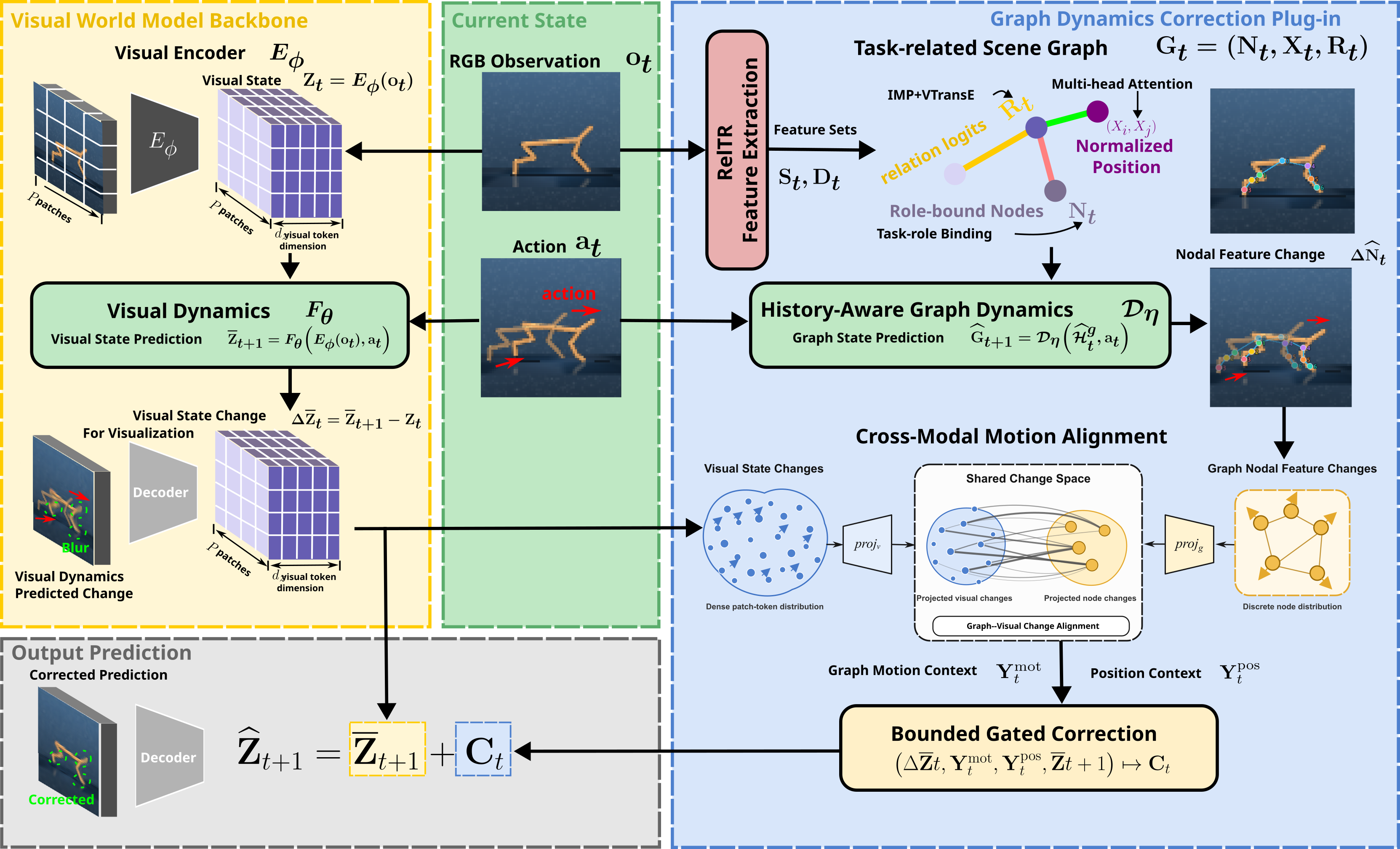}
    \caption{
    \textbf{Overview of JointGV.} Given the current RGB observation and action, the original visual world model (yellow) predicts the uncorrected next visual state $\overline{\mathbf{Z}}_{t+1}$ and its change $\Delta\overline{\mathbf{Z}}_t$. In parallel, JointGV (blue) uses Relation Transformer (RelTR) features to construct a task-conditioned scene graph and predicts the node-feature change $\Delta\widehat{\mathbf{N}}_t$. For the spatial-token case shown here, learned projections map the visual-token and graph-node changes into a shared space, where a change-alignment adapter produces the graph-motion context. The predicted node positions provide a complementary position context. A bounded, gated correction combines these contexts with the original visual prediction to produce $\widehat{\mathbf{Z}}_{t+1}=\overline{\mathbf{Z}}_{t+1}+\mathbf{C}_t$. Setting $\mathbf{C}_t=\mathbf{0}$ recovers the original prediction. Decoded images are shown only for visualization; the global-state adapter is described in Sec.~\ref{sec:motion_alignment}.
    }
\label{fig:jointgv_overview}
\vspace{-11pt}
\end{figure*}

JointGV is based on a cross-granularity dynamics hypothesis: although graph nodes and visual states share neither elements nor a feature space, their action-induced changes describe the same physical transition. JointGV couples these changes and combines the resulting graph-motion context with graph-predicted node positions to compute a bounded correction $\mathbf{C}_t$:
\begin{equation}
\widehat{\mathbf{Z}}_{t+1}
=
\overline{\mathbf{Z}}_{t+1}
+
\mathbf{C}_t.
\label{eq:corrected_prediction}
\end{equation}
Graph-learning and correction losses do not update the original visual encoder or dynamics. Figure~\ref{fig:jointgv_overview} shows the complete data flow.

\subsection{Task-Conditioned Scene Graph Construction}
\label{sec:graph_observer}

From RGB observation $\mathbf{o}_t$, JointGV constructs
\begin{equation}
\mathbf{G}_t
=
\left(
\mathbf{N}_t,
\mathbf{X}_t,
\mathbf{R}_t
\right),
\label{eq:scene_graph_definition}
\end{equation}
where $\mathbf{N}_t\in\mathbb{R}^{J\times d_g}$ contains $J$ task-role node features, $\mathbf{X}_t\in[-1,1]^{J\times2}$ contains their image positions, and $\mathbf{R}_t$ contains directed relation logits.

Learned task-role queries bind intermediate semantic and spatial RelTR features~\cite{cong2022reltr} to fixed graph slots. The resulting nodes exchange task-relevant information before a relation head predicts directed edges. Node identities, positions, visibility, and relations are training targets only; inference constructs the graph from RGB observations without simulator states. The feature binding, temporal query update, inter-node exchange, and relation head are detailed in Sec.~\ref{sec:supp_graph_observer}.

\subsection{History-Aware Graph Dynamics}
\label{sec:graph_dynamics}

Let \(H_g\geq1\) denote the number of most recent scene graphs supplied to the graph dynamics model. These graphs, together with the action \(\mathbf{a}_t\) applied between time \(t\) and \(t+1\), are used to predict the next scene graph. This design follows learned graph physics models that combine relation-aware message passing with short motion histories~\cite{sanchez2018graph,sanchez2020learning}. For node \(i\), we define its one-step change as \(\delta\mathbf{n}_{t,i}=\mathbf{n}_{t,i}-\mathbf{n}_{t-1,i}\). The current node feature and its changes over the history window are encoded as
\begin{equation}
\mathbf{g}^{(0)}_{t,i}
=
f_h\!\left(
\left[
\mathbf{n}_{t,i};
\delta\mathbf{n}_{t-H_g+2,i};
\ldots;
\delta\mathbf{n}_{t,i}
\right]
\right),
\label{eq:history_encoding}
\end{equation}
where \(f_h\) is a multilayer perceptron (MLP), and the superscript \((0)\) indicates the input to the first message-passing layer. When \(H_g=1\), the input contains only the current node feature \(\mathbf{n}_{t,i}\).

Before message passing, the relation logits \(\mathbf{R}_t\) are normalized over the \(K_r\) relation classes. At each of the \(L_G\) graph layers, every active node sends an action-conditioned message to the other active nodes. The relation probabilities of each node pair are embedded into its message, while the probability of the no-relation class controls how strongly that message contributes to the receiving node.

The final graph features are decoded into bounded node and relation-logit increments, denoted by \(\Delta\widehat{\mathbf{N}}_t\) and \(\Delta\widehat{\mathbf{R}}_t\), respectively. The next graph state is
\begin{equation}
\begin{aligned}
\widehat{\mathbf{N}}_{t+1}
&=
\mathbf{N}_t+\Delta\widehat{\mathbf{N}}_t,\\
\widehat{\mathbf{R}}_{t+1}
&=
\mathbf{R}_t+\Delta\widehat{\mathbf{R}}_t,
\end{aligned}
\label{eq:graph_increment_update}
\end{equation}
where each component of \(\Delta\widehat{\mathbf{N}}_t\) and \(\Delta\widehat{\mathbf{R}}_t\) is bounded by \(c_n>0\) and \(c_r>0\), respectively. The predicted node positions \(\widehat{\mathbf{X}}_{t+1}\) are decoded from \(\widehat{\mathbf{N}}_{t+1}\), giving the predicted scene graph \(\widehat{\mathbf{G}}_{t+1}=(\widehat{\mathbf{N}}_{t+1},\widehat{\mathbf{X}}_{t+1},\widehat{\mathbf{R}}_{t+1})\). The complete message-passing and increment-decoding operations are provided in Sec.~\ref{sec:supp_graph_dynamics}.

\paragraph{Self-fed dynamics training.}
One-step training always provides an observed graph as the input, whereas inference recursively uses previously predicted graphs. To reduce this mismatch, we train the graph dynamics through a \(K\)-step self-fed rollout, where \(K\geq1\) is the training horizon. Starting from a graph-history window constructed from observed image frames, rollout step \(k\in\{1,\ldots,K\}\) predicts
\begin{equation}
\begin{aligned}
\left(
\widehat{\mathbf{N}}_{t+k},
\widehat{\mathbf{R}}_{t+k}
\right)
=
\mathcal{D}_{\eta}\!\left(
\widehat{\mathcal{H}}^{g}_{t+k-1},
\mathbf{a}_{t+k-1}
\right),
\end{aligned}
\label{eq:graph_multistep_recursion}
\end{equation}
where \(\mathcal{D}_{\eta}\) is the graph dynamics model with parameters \(\eta\), \(\mathbf{a}_{t+k-1}\) is the action applied during the current transition, and \(\widehat{\mathcal{H}}^{g}_{t+k-1}\) contains the most recent \(H_g\) observed or predicted graphs. The first transition uses an observed graph history, after which every predicted graph is inserted into the history window for the following transition. Observed future graphs and simulator states are used only as supervision targets, not as inputs to the recursive rollout. The complete masked losses and the matched self-fed control for the original visual world model are provided in Sec.~\ref{sec:supp_training_objective}.

\subsection{Graph--Visual Correction Interface}
\label{sec:motion_alignment}

The interface is defined on predicted changes rather than absolute features. Graph nodes and visual states retain different information, but the changes induced by the same action isolate the physical transition represented by both predictions. The graph rollout then supplies two complementary cues: node changes describe how task-relevant entities evolve, while predicted node positions provide their spatial reference. This separation allows the same correction principle to operate on both spatial tokens and a global latent state.

JointGV operationalizes the cross-granularity dynamics hypothesis by coupling projected graph-node and visual-state changes. The two representations need not share elements, cardinality, or a feature space, and the interface is not tied to a particular coupling operator. We use Sinkhorn optimal transport for spatial tokens and normalized attention pooling for a global latent state.

\paragraph{Spatial-token alignment.}
For a spatial visual state, we instantiate the alignment adapter with Sinkhorn optimal transport (OT). Let \(\Delta\overline{\mathbf{z}}_{t,p}=[\Delta\overline{\mathbf{Z}}_t]_p\in\mathbb{R}^{d_z}\) denote the predicted change of visual token \(p\in\{1,\ldots,P\}\), and let \(\Delta\widehat{\mathbf{n}}_{t,i}=[\Delta\widehat{\mathbf{N}}_t]_i\in\mathbb{R}^{d_g}\) denote the predicted change of active graph node \(i\in\mathcal{I}\). Here, \(P\) is the number of visual tokens, \(d_z\) and \(d_g\) are the visual-token and graph-node feature dimensions, respectively, and \(\mathcal{I}\) is the set of active task-role-bound nodes.

Because the two changes belong to different feature spaces, $\operatorname{proj}_v:\mathbb{R}^{d_z}\rightarrow\mathbb{R}^{d_c}$ and $\operatorname{proj}_g:\mathbb{R}^{d_g}\rightarrow\mathbb{R}^{d_c}$ project and $\ell_2$-normalize them in a shared $d_c$-dimensional space. The visual-token and graph-node sets also have different sizes and no fixed one-to-one pairing. We use entropy-regularized OT for soft many-to-few change coupling; cosine attention provides an alternative learned coupling and is evaluated in Table~\ref{tab:dinowm_results}. The OT matching cost and transport plan are
\begin{equation}
\begin{aligned}
M_{t,pi}
&=
\left\|
\operatorname{proj}_v
\left(
\Delta\overline{\mathbf{z}}_{t,p}
\right)
-
\operatorname{proj}_g
\left(
\Delta\widehat{\mathbf{n}}_{t,i}
\right)
\right\|_2^2,\\
\boldsymbol{\Pi}_t^\epsilon
&=
\operatorname{Sinkhorn}_{\epsilon}
\left(
\mathbf{M}_t
\right),
\end{aligned}
\label{eq:motion_transport}
\end{equation}
where $\mathbf{M}_t\in\mathbb{R}^{P\times|\mathcal{I}|}$ contains the matching costs between all visual tokens and active graph nodes, $\epsilon>0$ is the entropic regularization coefficient, and $\boldsymbol{\Pi}_t^\epsilon\in\mathbb{R}^{P\times|\mathcal{I}|}$ is the resulting transport plan~\cite{cuturi2013sinkhorn}. We train the two projections using the standard debiased Sinkhorn
divergence $\mathcal{L}_{\mathrm{align}}$~\cite{feydy2019interpolating};
implementation details are provided in the supplementary material.

The same transport plan assigns a graph-motion context to each visual token:
\begin{equation}
\mathbf{y}^{\mathrm{mot}}_{t,p}
=
\sum_{i\in\mathcal{I}}
\widetilde{\Pi}_{t,pi}
\tanh\!\left(
\mathbf{W}_{gv}
\Delta\widehat{\mathbf{n}}_{t,i}
\right)
\in\mathbb{R}^{d_z}.
\label{eq:transported_graph_context}
\end{equation}
where \(\widetilde{\Pi}_{t,pi}\) is the row-normalized transport weight and \(\mathbf{W}_{gv}\in\mathbb{R}^{d_z\times d_g}\) maps each graph-node change to the visual-token dimension. The resulting \(\mathbf{y}^{\mathrm{mot}}_{t,p}\) contains the relation-aware graph change assigned to visual token \(p\).

\paragraph{Position-aware graph context.}
Graph--visual change alignment captures action-induced state evolution in the shared feature space but not absolute image location. JointGV therefore uses the next node positions predicted by the graph rollout to route predicted node features to nearby visual tokens, providing the spatial reference required for correction without ground-truth positions at inference. Let \(\widehat{\mathbf{x}}_{t+1,i}\in[-1,1]^2\) be the predicted image position of node \(i\), and let \(\mathbf{u}_p\in[-1,1]^2\) be the fixed image-grid coordinate of visual token \(p\). The position-aware graph context is
\begin{equation}
\begin{aligned}
\mathbf{y}^{\mathrm{pos}}_{t,p}
={}&
\sum_{i\in\mathcal{I}}
\exp\!\left(
-\frac{
\left\|
\mathbf{u}_p-\widehat{\mathbf{x}}_{t+1,i}
\right\|_2^2
}{
2\sigma_{\mathrm{pos}}^2
}
\right)\\
&\quad\cdot
\tanh\!\left(
\mathbf{W}_{\mathrm{pos}}
\widehat{\mathbf{n}}_{t+1,i}
\right).
\end{aligned}
\label{eq:position_context}
\end{equation}
The exponential term assigns a large weight when token \(p\) is close to the predicted position of node \(i\). The projected node feature \(\tanh(\mathbf{W}_{\mathrm{pos}}\widehat{\mathbf{n}}_{t+1,i})\) is the graph information delivered to nearby tokens. Here, \(\sigma_{\mathrm{pos}}>0\) controls the spatial range over which each node influences nearby visual tokens, and \(\mathbf{W}_{\mathrm{pos}}\in\mathbb{R}^{d_z\times d_g}\) maps the predicted node feature to the visual-token dimension. 

\paragraph{Bounded visual correction.}

The motion-based and position-based graph contexts provide complementary information: \(\mathbf{y}^{\mathrm{mot}}_{t,p}\) identifies graph-node changes compatible with the visual-token change, whereas \(\mathbf{y}^{\mathrm{pos}}_{t,p}\) identifies graph nodes located near the visual token. For token \(p\), the correction module concatenates both contexts with the predicted visual change and the original next-token prediction:
\begin{equation}
\boldsymbol{\chi}_{t,p}
=
\left[
\Delta\overline{\mathbf{z}}_{t,p};
\mathbf{y}^{\mathrm{mot}}_{t,p};
\mathbf{y}^{\mathrm{pos}}_{t,p};
\overline{\mathbf{z}}_{t+1,p}
\right],
\label{eq:correction_features}
\end{equation}
where \(\boldsymbol{\chi}_{t,p}\) is the complete input feature of the correction module for visual token \(p\).

The bounded correction for visual token \(p\) is
\begin{equation}
\mathbf{c}_{t,p}
=
s_c\,
\operatorname{sigmoid}\!\left(
f_{\gamma}(\boldsymbol{\chi}_{t,p})
\right)
\odot
\left[
c_{\max}
\tanh\!\left(
\frac{
f_c(\boldsymbol{\chi}_{t,p})
}{
c_{\max}
}
\right)
\right].
\label{eq:bounded_correction}
\end{equation}
Here, \(f_c\) predicts the correction, \(f_{\gamma}\) predicts its token-wise gate, \(\odot\) denotes element-wise multiplication, \(s_c\in[0,1]\) controls the overall correction strength, and \(c_{\max}>0\) bounds each component of the correction. Stacking the corrections of all \(P\) visual tokens gives \(\mathbf{C}_t=[\mathbf{c}_{t,1},\ldots,\mathbf{c}_{t,P}]^\top\), which is applied to the original visual prediction through Eq.~\eqref{eq:corrected_prediction}.

\paragraph{Global-vector alignment.}
When the visual state is a single global vector, e.g., in DreamerV3, there is no spatial token distribution on which to apply OT. In this case, $\operatorname{proj}_v$ denotes a separately learned projection from the global-state dimension to the same alignment dimension $d_c$. JointGV uses the predicted global visual-state change $\Delta\overline{\mathbf{Z}}_t$ to attend to graph-node changes. The attention weight of active node $i$ is
\begin{equation}
\alpha_{t,i}
=
\underset{i\in\mathcal{I}}{\operatorname{softmax}}
\left(
\frac{
\operatorname{proj}_v(\Delta\overline{\mathbf{Z}}_t)^\top
\operatorname{proj}_g(\Delta\widehat{\mathbf{n}}_{t,i})
}{
\tau
}
\right),
\label{eq:global_attention}
\end{equation}
where \(\tau>0\) is the attention temperature. The pooled graph-node change and its alignment loss are
\begin{equation}
\begin{aligned}
\Delta\widehat{\mathbf{n}}^{\mathrm{pool}}_t
&=
\sum_{i\in\mathcal{I}}
\alpha_{t,i}
\Delta\widehat{\mathbf{n}}_{t,i},\\
\mathcal{L}_{\mathrm{global}}
&=
\frac{1}{2}
\left\|
\operatorname{proj}_v(\Delta\overline{\mathbf{Z}}_t)
-
\operatorname{proj}_g
\left(
\Delta\widehat{\mathbf{n}}^{\mathrm{pool}}_t
\right)
\right\|_2^2.
\end{aligned}
\label{eq:global_alignment}
\end{equation}
The pooled graph change is then used by the same bounded correction principle. For global correction, the predicted nodes' projected features and two-dimensional positions are concatenated in fixed task-role order to form a global node context. This context and the pooled graph change are both supplied to the bounded, gated correction module. This adapter uses attention pooling followed by normalized half-squared \(\ell_2\) alignment; it does not perform a singleton Sinkhorn computation.

\subsection{Training Objective}
\label{sec:training_objective}

JointGV trains graph observation, graph dynamics, graph--visual alignment, and bounded correction while retaining the original visual-model objective. The uncorrected visual prediction is detached before alignment and correction, so these losses do not update the original visual encoder or dynamics. During recursive correction training, the visual model and graph dynamics are evaluated without gradients, and only the correction module is updated. The complete objective and recursive losses are provided in Sec.~\ref{sec:supp_training_objective}.

%% file: sec/Experiment_Result.tex
\begin{figure*}[t]
    \centering
    \includegraphics[width=0.8\textwidth]{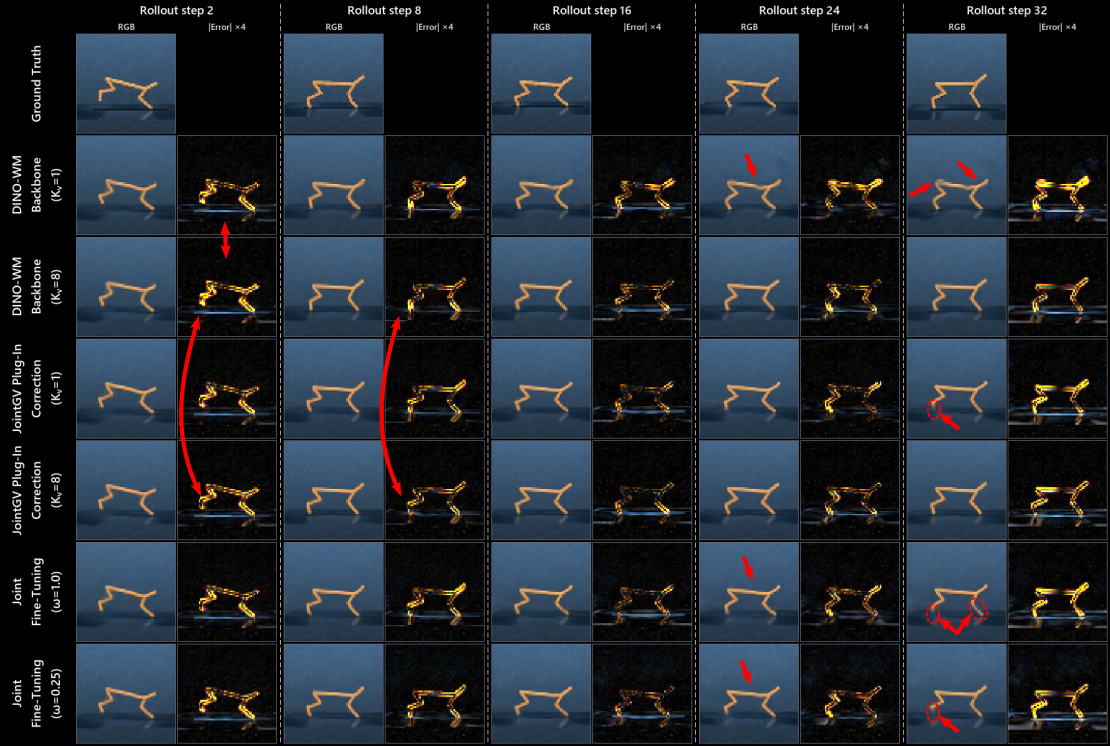}
    \vspace{-6pt} 
    \caption{\textbf{Qualitative comparison of DINO-WM and JointGV on a held-out DMC Cheetah Run sequence.} Columns show rollout steps 2, 8, 16, 24, and 32; adjacent maps show absolute RGB error amplified by four using the shared decoder. Red arrows mark representative structural differences. $K_v$ denotes the self-fed visual training horizon and $\omega$ the propagated correction-gradient fraction. Additional episodes and videos are provided in Sec.~\ref{sec:supp_qualitative}.}
\label{fig:dinowm_qualitative}
\vspace{-10pt}
\end{figure*}

\begin{table*}[t]
  \centering
  \footnotesize
  \setlength{\tabcolsep}{2.7pt}
  \renewcommand{\arraystretch}{0.91}
  \begin{tabular}{@{}lcccccccc@{}}
    \toprule
     & & \multicolumn{3}{c}{Joint-position error (px) $\downarrow$}
     & Token MSE $\downarrow$
     & RGB MSE$\times10^{3}$ $\downarrow$
     & SSIM $\uparrow$
     & LPIPS $\downarrow$ \\
    \cmidrule(lr){3-5}
    \cmidrule(lr){6-6}
    \cmidrule(lr){7-7}
    \cmidrule(lr){8-8}
    \cmidrule(lr){9-9}
    Method & Yr.
     & $H{=}8$ & $H{=}16$ & $H{=}32$
     & $H{=}32$ & $H{=}32$ & $H{=}32$ & $H{=}32$ \\
    \midrule

    \multicolumn{9}{@{}l}{\emph{Longer visual training}} \\
    DINO-WM ($K_v{=}1$)~\cite{zhou2024dinowm}
     & 2025 & 0.830 & 0.931 & 1.036 & 0.456 & 2.31 & 0.841 & 0.0435 \\
    DINO-WM + JointGV ($K_v{=}1$)
     & Ours & \textbf{0.730} & \textbf{0.796} & \textbf{0.852}
     & 0.341 & 1.84 & 0.861 & \textbf{0.0351} \\
    DINO-WM ($K_v{=}8$)
     & 2025 & 0.882 & 0.940 & 0.957 & 0.339 & 2.13 & 0.853 & 0.0420 \\
    DINO-WM + JointGV ($K_v{=}8$)
     & Ours & 0.750 & 0.815 & 0.863
     & \textbf{0.314} & \textbf{1.80} & \textbf{0.863} & 0.0352 \\

    \midrule
    \multicolumn{9}{@{}l}{\emph{Residual correction without a predicted graph}} \\
    Matched DINO-WM backbone
     & 2025 & 0.765 & 0.847 & 0.898 & 0.377 & 2.06 & 0.853 & 0.0366 \\
    Visual-only residual$^{*}$~\cite{lanier2026redraw,soni2026sandwich}
     & 2026 & 0.799 & 0.860 & 0.893 & 0.343 & 2.01 & 0.856 & 0.0380 \\
    Visual-only residual + node supervision\textsuperscript{*\S}
     & 2026 & 0.771 & 0.836 & 0.883 & 0.351 & 1.99 & 0.856 & 0.0362 \\

    \midrule
    \multicolumn{9}{@{}l}{\emph{Correction using a predicted graph rollout}} \\
    JointGV: mean graph-motion broadcast
     & Ours & 0.700 & 0.765 & 0.815 & 0.322 & 1.73 & 0.866 & 0.0326 \\
    JointGV: cosine change coupling$^\dagger$
     & Ours & \textbf{0.669} & \textbf{0.733} & 0.793
     & \textbf{0.304} & 1.69 & \textbf{0.869} & 0.0322 \\
    JointGV: OT change coupling
     & Ours & 0.693 & 0.750 & \textbf{0.791}
     & \textbf{0.304} & \textbf{1.67} & \textbf{0.869}
     & \textbf{0.0316} \\

    \bottomrule
  \end{tabular}
  \vspace{-6pt}
  \caption{
    \textbf{DINO-WM results on 280 matched DMC Cheetah Run test windows.}
    Values at horizon $H$ average steps 1 through $H$; Token MSE denotes DINOv2-token MSE. Rows are single runs and are compared only within matched blocks. $^{*}$The visual-residual controls adapt residual-correction principles rather than reproduce the cited methods. \textsuperscript{\S}The annotation-matched control uses the same auxiliary node targets but receives no predicted graph states or graph features. $^\dagger$The cosine run has its own matched backbone. Three-seed results and additional ablations are provided in Sec.~\ref{sec:supp_ablation}.
  }
  \label{tab:dinowm_results}
  \vspace{-13pt}
\end{table*}

\begin{figure*}[t]
    \centering
    \includegraphics[width=0.85\textwidth]{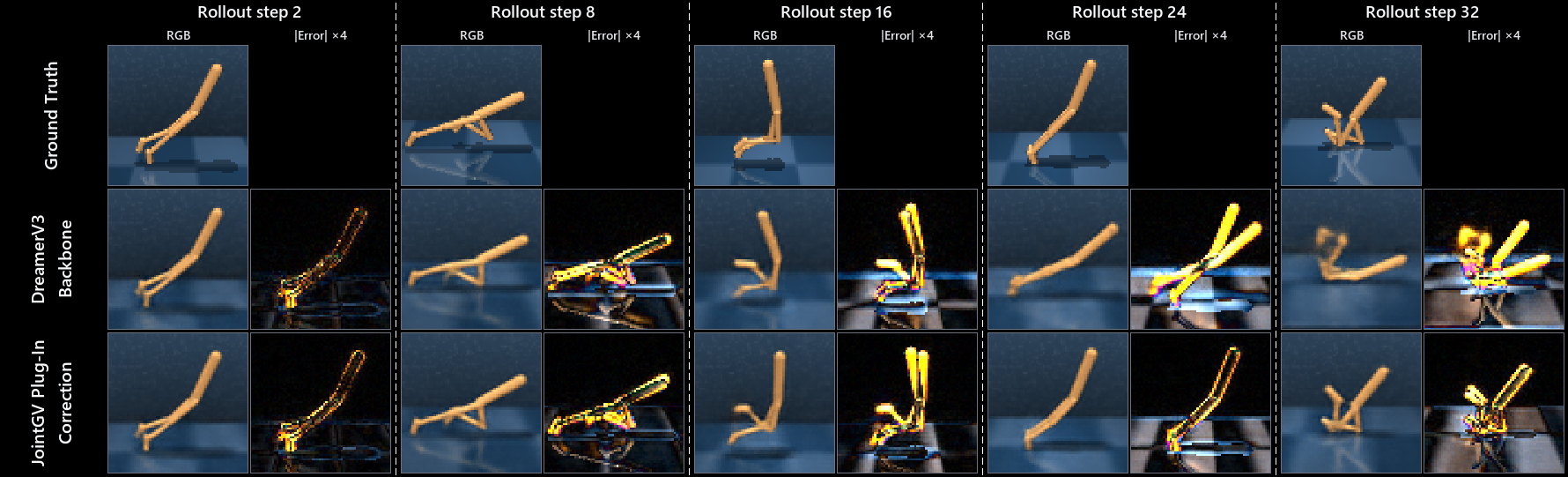}
    \caption{\textbf{Qualitative comparison on a held-out DMC Walker Walk validation sequence.} Columns show rollout steps 2, 8, 16, 24, and 32, with absolute RGB error amplified by a factor of four. JointGV uses the DreamerV3 global-state change to pool the graph-node changes and applies the result as a bounded global correction. The corrected rollout better preserves the articulated pose at later steps.}
\label{fig:dreamer_qualitative}
\vspace{-3pt}
\end{figure*}

\begin{table*}[t]
  \centering
  \footnotesize
  \setlength{\tabcolsep}{5pt}
  \renewcommand{\arraystretch}{0.95}
  \begin{tabular}{@{}lllccc@{}}
    \toprule
    Visual world model & Visual state & Task
    & Original
    & + JointGV
    & Reduction (\%) $\uparrow$ \\
    \midrule
    DINO-WM & Spatial tokens & Cheetah Run
    & $0.942\pm0.046$ & $\mathbf{0.790\pm0.011}$ & $\mathbf{16.0\pm3.7}$ \\
    DreamerV3 & Global latent & Cheetah Run
    & $0.467\pm0.009$ & $\mathbf{0.415\pm0.009}$ & $\mathbf{11.1\pm2.6}$ \\
    DINO-WM & Spatial tokens & Walker Walk
    & $10.740\pm0.922$ & $\mathbf{7.682\pm0.151}$ & $\mathbf{28.0\pm7.8}$ \\
    DreamerV3 & Global latent & Walker Walk
    & $5.835\pm0.093$ & $\mathbf{4.336\pm0.019}$ & $\mathbf{25.7\pm1.3}$ \\
    \bottomrule
  \end{tabular}
  \caption{
    \textbf{JointGV across visual states and control tasks.}
    Values are average joint-position errors over rollout steps 1--32, reported as mean and standard deviation across three independently trained seeds. Every model--task pairing improves in all three seeds. Per-horizon and visual metrics are provided in Sec.~\ref{sec:supp_ablation}.
  }
  \label{tab:dreamerv3_results}
  \vspace{-7pt}
\end{table*}

\section{Experiments and Results}
\label{sec:experiments}

\subsection{Experimental Setup}
\label{sec:experimental_setup}

We evaluate DINO-WM and DreamerV3 on DMC Cheetah Run and Walker Walk across three seeds, and extend the DMC evaluation to contact-rich manipulation with Finger Turn Hard~\cite{tassa2018deepmind}. Tables~\ref{tab:dinowm_results}--\ref{tab:finger_results} report DINO-WM on Cheetah Run, all four locomotion pairings over three seeds, and Finger Turn Hard; complete metrics are in Sec.~\ref{sec:supp_ablation}. For DreamerV3, we reproduce its online actor--critic training, then train the backbone and JointGV on the same fixed dataset collected using the reproduced policy and random actions. This avoids exposing separately trained prediction models to different transition distributions~\cite{hafner2025mastering}; details are provided in Sec.~\ref{sec:supp_evaluation}. We additionally evaluate DINO-WM, DreamerV3, and LeWorldModel on recordings of a physical inverted-pendulum system built with an RM65-B arm~\cite{realman_rm65b} and a rotary gear damper~\cite{toyou_trd_cg5da}, using the same observed image history and recorded actions.

We report two-dimensional joint-position error, RGB mean squared error (MSE), structural similarity (SSIM)~\cite{wang2004image}, Learned Perceptual Image Patch Similarity (LPIPS)~\cite{zhang2018unreasonable}, and DINOv2-token MSE for DINO-WM. Values at horizon $H$ average rollout steps 1 through $H$. Joint positions are measured by a frozen RGB observer separate from JointGV's graph branch. We also rescore the predictions with an independently trained heatmap-based keypoint detector to avoid dependence on the RelTR-based evaluator. Metric definitions and protocols are provided in Secs.~\ref{sec:supp_evaluation} and~\ref{sec:supp_observer}.

For repeated experiments, we report the mean and standard deviation across three training seeds; within each seed, 95\% confidence intervals use 2,000 paired bootstrap resamples of the fixed test windows.

Controls compare longer self-fed visual training, visual-only residuals inspired by ReDRAW and Sandwich-Residuals~\cite{lanier2026redraw,soni2026sandwich}, mean graph-motion broadcasting, cosine and OT change coupling, and correction-gradient propagation. The residual controls adapt selected principles rather than reproduce the complete methods. Simulator annotations, and sensor-derived geometric node targets for the physical data, supervise graph observation and dynamics during training but are not supplied during recursive inference. The annotation-matched visual residual receives the same auxiliary node targets but no predicted graph states or graph features; its results are reported in Sec.~\ref{sec:supp_ablation}. Actors trained with original or JointGV-corrected DINO-WM rollouts are compared in a matched MuJoCo model (Table~\ref{tab:control}).

\subsection{Long-Horizon Prediction with DINO-WM}
\label{sec:dinowm_results}

Table~\ref{tab:dinowm_results} and Fig.~\ref{fig:dinowm_qualitative} show that JointGV reduces joint-position error across all reported horizons, improves the latent and decoded-image metrics, and better preserves articulated shape during rollout.

Longer self-fed visual training improves DINO-WM at later horizons but worsens its shorter-horizon joint-position error. JointGV improves both the $K_v=1$ and $K_v=8$ backbones, showing that its gain is not explained by longer visual training alone.

\textbf{Contact-rich manipulation.} On DMC Finger Turn Hard, JointGV reduces 32-step joint-position error by 27.6\% and RGB MSE by 51.8\%, while better preserving the finger--spinner configuration over the rollout (Table~\ref{tab:finger_results} and Fig.~\ref{fig:supp_finger_rollout}). Additional metrics are provided in Sec.~\ref{sec:supp_ablation}.

\begin{table}[t]
  \centering
  \footnotesize
  \setlength{\tabcolsep}{4pt}
  \renewcommand{\arraystretch}{0.95}
  \begin{tabular}{@{}lccc@{}}
    \toprule
     & \multicolumn{3}{c}{Joint-position error (px) $\downarrow$} \\
    \cmidrule(lr){2-4}
    Method & $H{=}8$ & $H{=}16$ & $H{=}32$ \\
    \midrule
    DINO-WM backbone~\cite{zhou2024dinowm}
      & 10.813 & 12.476 & 14.861 \\
    DINO-WM + JointGV
      & \textbf{9.444} & \textbf{9.911} & \textbf{10.754} \\
    \bottomrule
  \end{tabular}
  \vspace{-6pt}
  \caption{\textbf{DINO-WM on 280 DMC Finger Turn Hard test windows.} Values at $H$ average steps 1 through $H$; additional metrics are provided in Sec.~\ref{sec:supp_ablation}.}
  \label{tab:finger_results}
  \vspace{-14pt}
\end{table}

\subsection{Evidence for Graph--Visual Change Coupling}

Table~\ref{tab:dinowm_results} first separates graph-conditioned correction from residual capacity and additional supervision. The visual-only residual changes the 32-step joint-position error only from $0.898$ to $0.893$ pixels, whereas JointGV reduces it to $0.791$ pixels. Across three seeds, the corresponding reductions are $4.3\%\pm4.3\%$ and $16.0\%\pm3.7\%$. Supplying the residual with the same node annotations, but no predicted graph, reaches only $0.883$ pixels. The larger gain therefore comes from conditioning the correction on the predicted graph rollout, rather than from residual capacity or matched node supervision alone.

Mean graph-motion broadcasting improves the matched visual backbone, showing that the predicted graph rollout already provides a useful correction signal without token-specific routing. Cosine and OT coupling further reduce the 32-step error to $0.793$ and $0.791$ pixels, respectively. Their similar accuracy shows that the correction interface is not tied to one operator. The additional gain over broadcasting is largest when non-adjacent parts approach in image space, where assigning differentiated graph motion becomes more important. An independently trained heatmap-based keypoint network also measures a $13.6\%$ reduction and finds improvement on $95.4\%$ of the test windows. Complete paired comparisons and confidence intervals, and stratified analyses are provided in Secs.~\ref{sec:supp_ablation} and~\ref{sec:supp_stratified}.

\subsection{Portability to a Global Latent World Model}
\label{sec:dreamer_results}

Table~\ref{tab:dreamerv3_results} evaluates JointGV with spatial-token and global-latent states on Cheetah Run and Walker Walk. It improves every model--task pairing in all three seeds, reducing average 32-step joint-position error by $11.1$--$28.0\%$. For DreamerV3, graph-node changes are pooled using the global visual-state change rather than spatial-token OT, showing that the correction principle does not depend on a visual-token grid. Figure~\ref{fig:dreamer_qualitative} shows the corresponding Walker Walk rollout.

\begin{figure}[t]
\centering
\includegraphics[width=0.92\linewidth]{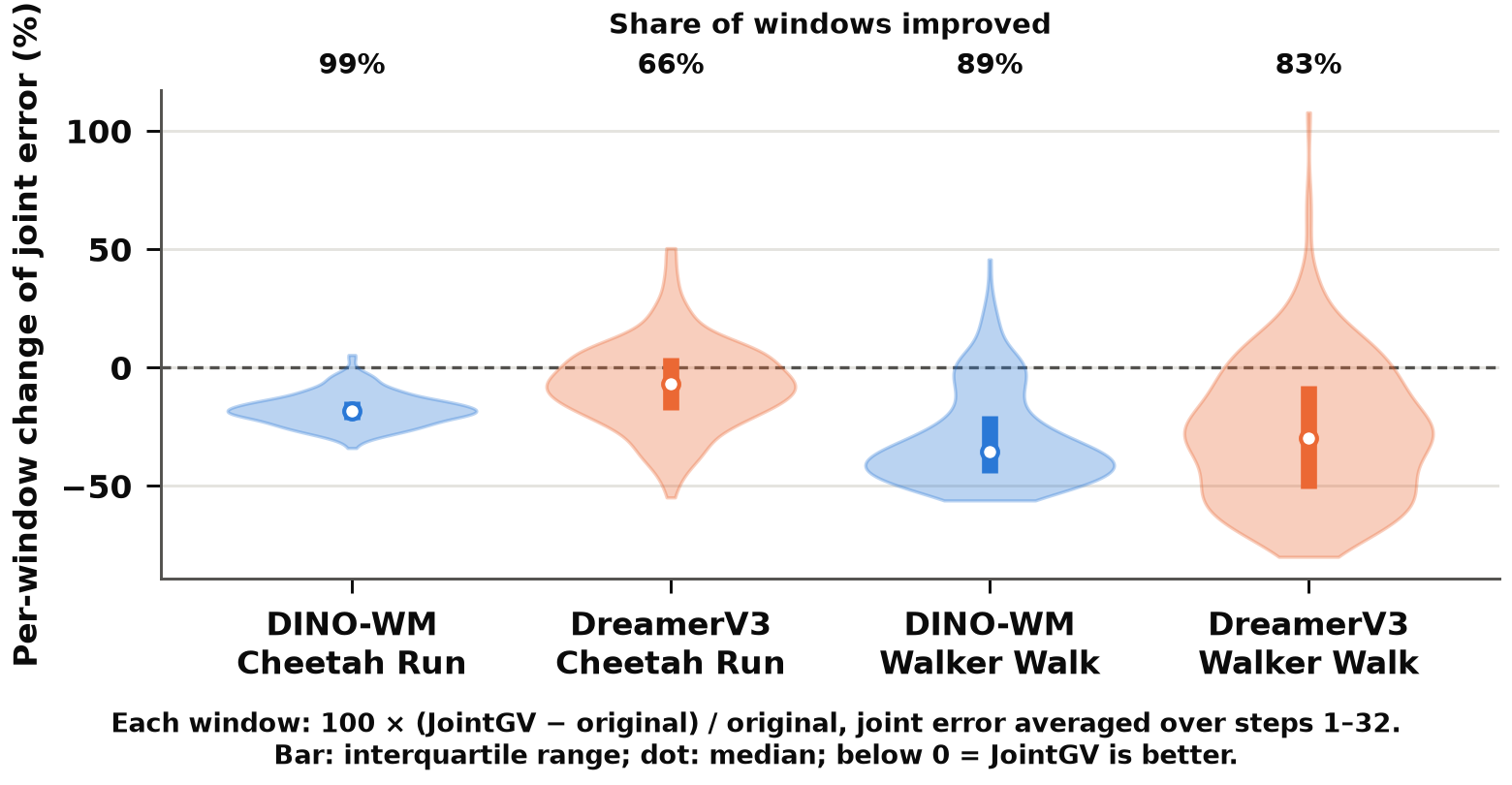}
\vspace{-6pt}
\caption{\textbf{Per-window effect across four model--task pairings.}
Each value is the relative change in 32-step average joint-position error from the original prediction to JointGV; values below zero indicate improvement. JointGV improves 99\%, 66\%, 89\%, and 83\% of the respective test windows, showing that the aggregate gains are not produced by a small subset of rollouts.}
\label{fig:supp_paired_windows}
\vspace{-7pt}
\end{figure}

The per-window distributions in Fig.~\ref{fig:supp_paired_windows} further show that these averages are not driven by a few favorable rollouts. JointGV improves 277 of 280 DINO-WM Cheetah Run windows, 186 of 280 DreamerV3 Cheetah Run windows, 250 of 280 DINO-WM Walker Walk windows, and 232 of 280 DreamerV3 Walker Walk windows. The broader DreamerV3 distributions also show that the correction magnitude remains task dependent, despite a consistent mean improvement across all three seeds.

\subsection{Inverted-Pendulum Prediction and Control}
\label{sec:real_robot}

\textbf{Real-robot prediction.}
On 156 matched windows from real-robot sequences, JointGV reduces DINO-WM's average 32-step pole-angle error from $12.56^\circ$ to $6.69^\circ$ and its pole-angle-change error from $3.09^\circ$ to $1.46^\circ$, obtaining the lower change error on $82.1\%$ of the windows. DreamerV3 and LeWorldModel~\cite{maes2026leworldmodel} also improve.

\textbf{Simulated control.}
We test whether the prediction gains support control in a MuJoCo model of the same RM65-B inverted-pendulum system. Two actors are trained in imagination with identical action constraints and a shared reward predictor; they differ only in using original or JointGV-corrected DINO-WM rollouts. On 50 unseen initial states, the JointGV-rollout actor completes 46 six-second episodes, compared with none for the original-rollout actor, and its mean time before termination rises from $2.28$ to $5.86$\,s (Table~\ref{tab:control}).

\begin{figure}[t]
\centering
\includegraphics[width=0.9\linewidth]{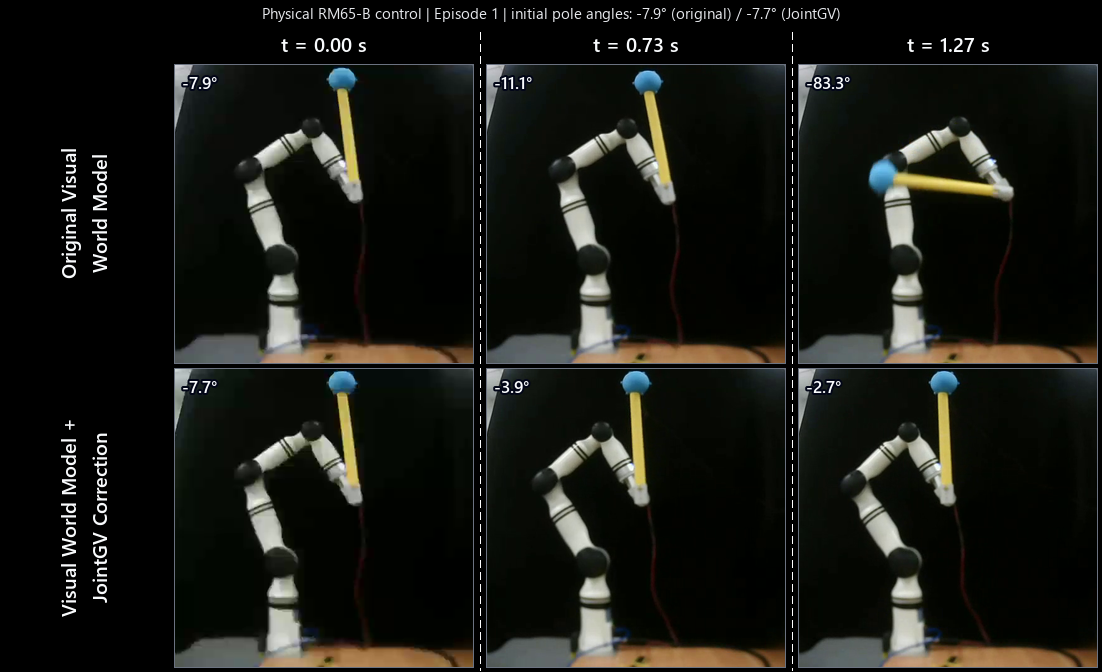}
\caption{\textbf{Preliminary physical RM65-B control trial.}
The two actors start from comparable pole angles. At $1.27$ seconds, the pole controlled using the original visual-world-model rollout has fallen to $-83.3^\circ$, whereas the actor using the JointGV-corrected rollout keeps it near upright at $-2.7^\circ$. This trial demonstrates short-duration physical deployment rather than stable six-second balancing.}
\label{fig:robot_control}
\vspace{-15pt}
\end{figure}

\textbf{Physical trial.}
Figure~\ref{fig:robot_control} shows a preliminary physical control trial from comparable initial pole angles. After $1.27$ seconds, the original-rollout actor reaches $-83.3^\circ$, whereas the JointGV-rollout actor remains at $-2.7^\circ$. Stable six-second physical balancing has not yet been achieved because high-speed operation introduces non-negligible command delay. We therefore report this trial as a deployment diagnostic and Table~\ref{tab:control} as the aggregate control evaluation; further details are provided in Sec.~\ref{sec:supp_robot_simulation}.

% \begin{figure}[t]
% \centering
% \includegraphics[width=\linewidth]{figures/cartpole_actor_seed42003_t0-1-2-3-4-5s.png}
% \caption{\textbf{MuJoCo inverted-pendulum control.} From the same unseen initial state, the actor trained with original DINO-WM rollouts falls after 1.44 seconds, whereas the actor trained with JointGV-corrected rollouts completes the six-second horizon.}
% \label{fig:robot_control}
% \vspace{-15pt}
% \end{figure}
\begin{table}[t]
\centering
\footnotesize
\setlength{\tabcolsep}{4pt}
\begin{tabular}{@{}lcc@{}}
\toprule
Measure (50 unseen initial states) & Original & + JointGV \\
\midrule
Six-second completion $\uparrow$ & 0/50 & \textbf{46/50} \\
Mean time before termination (s) $\uparrow$ & 2.28 & \textbf{5.86} \\
Shared reward-head MSE $\downarrow$ & 0.0851 & \textbf{0.0375} \\
\bottomrule
\end{tabular}
\vspace{-6pt}
\caption{\textbf{MuJoCo inverted-pendulum control.} Actors are trained with original or JointGV-corrected DINO-WM rollouts; reward-head MSE is a diagnostic.}
\label{tab:control}
\vspace{-15pt}
\end{table}

\section{Discussion and Conclusion}
\label{sec:discussion_conclusion}

JointGV improves long-horizon prediction across spatial-token and global-latent visual models, DMC articulated locomotion,contact-rich DMC manipulation, and real-robot sequences. These results support the central cross-granularity hypothesis: graph nodes and visual states without shared elements can still be coupled through changes induced by the same action. The stronger coupling advantage when non-adjacent parts approach further identifies where learned graph--visual correspondence is most useful.

The coupling analysis clarifies the role of graph--visual alignment. Cosine and OT coupling obtain similar overall accuracy, indicating that the correction interface is not tied to a particular operator. Both outperform broadcasting the same mean graph motion to every visual token, and this difference grows when non-adjacent parts approach in image space. The benefit therefore comes from associating visual changes with differentiated graph motion, particularly when the projected structure becomes ambiguous.

The visual-only and supervision-matched residual controls show that the gain is not explained by correction capacity or additional node annotations. Longer visual training also does not reproduce JointGV's short- and long-horizon improvements, while propagating correction gradients into the original visual dynamics degrades their native prediction. JointGV therefore uses the graph rollout as an external, gradient-isolated correction source rather than another branch of the original latent pathway.

JointGV adds computation and currently assumes supervised task-defined node roles. Its correction also depends on graph observation and prediction quality; overlapping links can cause node-localization errors. Automatic role discovery, confidence-aware correction, broader visual models, and stable physical closed-loop deployment remain future work. The current hardware system exhibits non-negligible control-command delay at high speed, preventing reliable six-second physical balancing despite the improved visual predictions.

Overall, JointGV provides a plug-in interface through which separately predicted relational dynamics correct long-horizon visual rollouts without restructuring the original visual world model.

%% file: sec/X_suppl.tex
\clearpage
\setcounter{page}{1}
\setcounter{section}{0}
\setcounter{figure}{0}
\setcounter{table}{0}
\setcounter{equation}{0}
\renewcommand{\thesection}{S\arabic{section}}
\renewcommand{\thesubsection}{\thesection.\arabic{subsection}}
\renewcommand{\thefigure}{S\arabic{figure}}
\renewcommand{\thetable}{S\arabic{table}}
\renewcommand{\theequation}{S\arabic{equation}}
\renewcommand{\theHsection}{supp.section.\arabic{section}}
\renewcommand{\theHsubsection}{supp.subsection.\arabic{section}.\arabic{subsection}}
\renewcommand{\theHfigure}{supp.figure.\arabic{figure}}
\renewcommand{\theHtable}{supp.table.\arabic{table}}
\renewcommand{\theHequation}{supp.equation.\arabic{equation}}
\maketitlesupplementary

\section*{LLM Usage Statement}

OpenAI ChatGPT/GPT and Anthropic Claude were used to assist with writing and editing, related-work and citation checking, experimental design and analysis, figures and visual materials, data preparation, and refinement of research ideas, claims, and conclusions. They were also used to generate and refine portions of the code and implementation. The full manuscript or major portions of it were provided to OpenAI ChatGPT/GPT for reviewer-style feedback. All tool outputs, references, implementations, experiments, results, and claims were reviewed and verified by the authors, who take full responsibility for the submitted work.

\section{Detailed Scene Graph Construction}
\label{sec:supp_graph_observer}

This section provides the implementation details omitted from Sec.~\ref{sec:graph_observer}. The observer constructs $\mathbf{G}_t=(\mathbf{N}_t,\mathbf{X}_t,\mathbf{R}_t)$ from the current RGB observation. Simulator states, per-frame object identities, and relation annotations are training targets only and are not supplied at inference.

\subsection{RelTR Features and Task-Role Binding}

Given RGB image $\mathbf{o}_t$, the pretrained Relation Transformer (RelTR)~\cite{cong2022reltr} provides sparse semantic features $\mathbf{S}_t\in\mathbb{R}^{M_s\times d_s}$ and dense spatial features $\mathbf{D}_t=[\mathbf{d}_{t,1},\ldots,\mathbf{d}_{t,M_d}]^\top\in\mathbb{R}^{M_d\times d_d}$. Here, $M_s$ and $M_d$ are the numbers of sparse and dense features, while $d_s$ and $d_d$ are their dimensions. Dense feature $\mathbf{d}_{t,m}$ has a fixed normalized image coordinate $\mathbf{u}_m\in[-1,1]^2$.

RelTR's final predicate predictions do not preserve the fixed task roles required by recursive graph prediction. We therefore retain its intermediate features and bind them to $J$ task-defined role slots. Role $i$ has a learned query $\mathbf{q}_i\in\mathbb{R}^{d_g}$, where $d_g$ is the node-feature dimension. At inference, the query locates its assigned role from RGB features without simulator states or per-frame object labels.

The sparse and dense attention operations follow standard multi-head and scaled dot-product attention~\cite{vaswani2017attention}. A first pass with $\mathbf{q}_i$ reads the current RGB features and produces an anchor feature $\mathbf{h}^{(0)}_{t,i}\in\mathbb{R}^{d_g}$. This feature determines how much information the query receives from the previous node.

\subsection{Bounded Temporal Query Correction}

Each role is located again from the current RGB frame. The previous node supplies a bounded hint when the role is briefly occluded. Let $\mathbf{n}^{\mathrm{LN}}_{t-1,i}=\operatorname{LN}(\mathbf{n}_{t-1,i})$ be the layer-normalized previous node. The history-conditioned query is
\begin{equation}
\begin{aligned}
\gamma^{o}_{t,i}
&=\operatorname{sigmoid}\!\left(
f_o\!\left[
\mathbf{h}^{(0)}_{t,i};
\mathbf{n}^{\mathrm{LN}}_{t-1,i}
\right]\right),\\
\widetilde{\mathbf{q}}_{t,i}
&=\mathbf{q}_i+s_oc_o\gamma^{o}_{t,i}
\tanh\!\left(
\mathbf{W}_h\mathbf{n}^{\mathrm{LN}}_{t-1,i}
\right).
\end{aligned}
\label{eq:supp_temporal_query}
\end{equation}
Here, the superscript $o$ identifies the observer gate, $f_o$ predicts its scalar value, $\mathbf{W}_h\in\mathbb{R}^{d_g\times d_g}$ projects the previous node, $s_o\in[0,1]$ controls the historical contribution, and $c_o>0$ bounds it. Setting $s_o=0$ recovers independent per-frame role binding.

The history-conditioned query binds the RelTR features to task role $i$:
\begin{equation}
\mathbf{h}_{t,i}
=
\operatorname{MHA}\!\left(
\widetilde{\mathbf{q}}_{t,i},
\mathbf{S}_t,
\mathbf{S}_t
\right)
+
\operatorname{Attn}\!\left(
\widetilde{\mathbf{q}}_{t,i},
\mathbf{D}_t,
\mathbf{D}_t
\right),
\label{eq:role_binding}
\end{equation}
where the two terms provide semantic and spatial evidence, respectively.

The updated query reads the sparse and dense RelTR features again. Adding their attention outputs gives role feature $\mathbf{h}_{t,i}$. Let $A_{t,i,m}$ be its normalized dense-attention weight at location $m$. The node position is
\begin{equation}
\mathbf{x}_{t,i}
=
\operatorname{clip}\!\left(
\sum_{m=1}^{M_d}A_{t,i,m}\mathbf{u}_m
+c_x\tanh\!\left(f_x(\mathbf{h}_{t,i})\right),
-1,1
\right),
\label{eq:supp_node_position}
\end{equation}
where $f_x:\mathbb{R}^{d_g}\rightarrow\mathbb{R}^{2}$ predicts a position offset and $c_x>0$ bounds its magnitude.

\subsection{Inter-Node Exchange and Relation Logits}

The role queries initially read the image independently. We then apply one round of Iterative Message Passing (IMP)~\cite{xu2017scenegraph}, allowing each active node to receive information from the other task-relevant nodes. For an ordered pair with subject node $s$ and object node $o$, the initial pair feature is
\begin{equation}
\mathbf{e}^{0}_{t,s,o}
=
f_e\!\left[
\mathbf{h}_{t,s};
\mathbf{h}_{t,o};
\mathbf{x}_{t,o}-\mathbf{x}_{t,s}
\right],
\label{eq:supp_pair_feature}
\end{equation}
where $f_e$ is a learned multilayer perceptron (MLP). A shared MLP sends message $\mathbf{m}_{t,s\rightarrow o}=f_m([\mathbf{h}_{t,s};\mathbf{e}^{0}_{t,s,o}])$. The received messages update node $o$:
\begin{equation}
\begin{aligned}
\overline{\mathbf{m}}_{t,o}
&=
\frac{1}{\max(J_a-1,1)}
\sum_{\substack{s\in\mathcal I\\s\neq o}}
\mathbf{m}_{t,s\rightarrow o},\\
\mathbf{n}_{t,o}
&=
\operatorname{LN}\!\left[
\operatorname{GRU}\!\left(
\overline{\mathbf{m}}_{t,o},
\mathbf{h}_{t,o}
\right)\right].
\end{aligned}
\label{eq:supp_imp_update}
\end{equation}
Here, $\mathcal I$ is the active-node set, $J_a=|\mathcal I|$ is its size, $\operatorname{GRU}$ is a gated recurrent unit, and $\operatorname{LN}$ is layer normalization.

IMP updates node features but does not assign semantic relations. For relation class $k$, we use the VTransE-inspired translation score
\begin{equation}
T_k\!\left(\mathbf{n}_{t,s},\mathbf{n}_{t,o}\right)
=
-\frac{1}{d_g}
\left\|
\mathbf{W}_s\mathbf{n}_{t,s}
+
\boldsymbol{\rho}_k
-
\mathbf{W}_o\mathbf{n}_{t,o}
\right\|_2^2,
\label{eq:vtranse_score}
\end{equation}
where $\mathbf{W}_s$ and $\mathbf{W}_o$ project the subject and object features and $\boldsymbol{\rho}_k$ is the learned translation for relation class $k$. We add a learned pair-specific residual:
\begin{equation}
\begin{aligned}
\mathbf{r}^{\mathrm{res}}
&=
f_r\!\left[
\mathbf{n}_{t,s};
\mathbf{n}_{t,o};
\mathbf{e}^{0}_{t,s,o};
\mathbf{x}_{t,o}-\mathbf{x}_{t,s}
\right],\\
[\mathbf{R}_t]_{s,o,k}
&=
T_k(\mathbf{n}_{t,s},\mathbf{n}_{t,o})
+
[\mathbf{r}^{\mathrm{res}}]_k.
\end{aligned}
\label{eq:supp_relation_logit}
\end{equation}
The shared MLP $f_r$ outputs one residual for each of the $K_r$ relation classes. Softmax over $k$ gives the relation distribution for this ordered pair; class $k=0$ denotes no relation.

\section{Detailed History-Aware Graph Dynamics}
\label{sec:supp_graph_dynamics}

This section expands the message passing and bounded graph update summarized in Sec.~\ref{sec:graph_dynamics}. The history input is defined in Eq.~\eqref{eq:history_encoding}.

\subsection{Relation-Aware Message Passing}

For ordered pair $(s,o)$, let $\mathbf{p}_{t,s,o}=\operatorname{softmax}([\mathbf{R}_t]_{s,o,:})$ be its distribution over the $K_r$ relation classes. We set $\beta_{t,s,o}=1-[\mathbf{p}_{t,s,o}]_0$ and $\mathbf{e}^{(\ell)}_{t,s,o}=\mathbf{W}^{(\ell)}_R\mathbf{p}_{t,s,o}$, where $\mathbf{W}^{(\ell)}_R:\mathbb{R}^{K_r}\rightarrow\mathbb{R}^{d_g}$ is learned at graph layer $\ell$. The no-relation probability controls message strength, while the projected distribution supplies the relation feature. The action-conditioned message and node update are
\begin{equation}
\begin{aligned}
\mathbf{m}^{(\ell)}_{t,s\rightarrow o}
&=
\beta_{t,s,o}f_m^{(\ell)}
\!\left[
\mathbf{g}^{(\ell)}_{t,s};
\mathbf{g}^{(\ell)}_{t,o};
\mathbf{e}^{(\ell)}_{t,s,o};
\mathbf{a}_t
\right],\\
\overline{\mathbf{m}}^{(\ell)}_{t,o}
&=
\frac{
\sum_{s\neq o}\mathbf{m}^{(\ell)}_{t,s\rightarrow o}
}{
\sum_{s\neq o}\beta_{t,s,o}+\varepsilon_e
},\\
\mathbf{g}^{(\ell+1)}_{t,o}
&=
\operatorname{LN}\!\left[
\mathbf{g}^{(\ell)}_{t,o}
+
f_u^{(\ell)}
\!\left[
\mathbf{g}^{(\ell)}_{t,o};
\overline{\mathbf{m}}^{(\ell)}_{t,o};
\mathbf{a}_t
\right]\right].
\end{aligned}
\label{eq:supp_dynamics_message}
\end{equation}
Here, $\varepsilon_e>0$ prevents division by zero, and $f_m^{(\ell)}$ and $f_u^{(\ell)}$ are learned MLPs. This update is repeated for $L_G$ graph layers.

\subsection{Bounded Graph Update and Self-Fed Rollout}

The final graph features predict bounded node and relation-logit increments:
\begin{equation}
\begin{aligned}
\Delta\widehat{\mathbf{n}}_{t,i}
&=
c_n\tanh\!\left(
f_n(\mathbf{g}^{(L_G)}_{t,i})
\right),\\
\Delta\widehat{\mathbf{r}}_{t,s,o}
&=
c_r\tanh\!\left(
f_R\!\left[
\mathbf{g}^{(L_G)}_{t,s};
\mathbf{g}^{(L_G)}_{t,o};
\mathbf{a}_t
\right]\right).
\end{aligned}
\label{eq:supp_graph_increments}
\end{equation}
Here, $f_n$ and $f_R$ decode the increments, while $c_n>0$ and $c_r>0$ bound each component. Stacking these outputs gives the updates in Eq.~\eqref{eq:graph_increment_update}. The position decoder $f_{\mathrm{pos}}:\mathbb{R}^{d_g}\rightarrow\mathbb{R}^{2}$ gives $\widehat{\mathbf{x}}_{t+1,i}=\tanh(f_{\mathrm{pos}}(\widehat{\mathbf{n}}_{t+1,i}))$.

The $K$-step training rollout follows Eq.~\eqref{eq:graph_multistep_recursion}. Its first prediction uses an observed graph-history window. Each later prediction uses the previous predicted graph instead of an observed future graph. Observed future graphs and simulator states serve only as training targets.

\section{Detailed Graph--Visual Alignment and Correction}
\label{sec:supp_alignment_correction}

This section provides implementation details omitted from Sec.~\ref{sec:motion_alignment}. Sinkhorn iterations follow Cuturi~\cite{cuturi2013sinkhorn}; we do not repeat their standard derivation.

\subsection{Spatial-Token Alignment}

The projection functions in Eq.~\eqref{eq:motion_transport} are learned bias-free linear layers followed by optional set centering and safe row-wise $\ell_2$ normalization. They map visual-token changes $\Delta\overline{\mathbf{z}}_{t,p}\in\mathbb{R}^{d_z}$ and graph-node changes $\Delta\widehat{\mathbf{n}}_{t,i}\in\mathbb{R}^{d_g}$ into the same $d_c$-dimensional space.

We use the debiased Sinkhorn divergence of Feydy et al.~\cite{feydy2019interpolating} as $\mathcal{L}_{\mathrm{align}}$ and clamp small negative numerical values to zero. The variance-floor loss $\mathcal{L}_{\mathrm{var}}$ penalizes a projected dimension when its sample standard deviation falls below $\gamma_{\sigma}>0$.

For each visual token, transport weights are normalized over active graph nodes before the motion context in Eq.~\eqref{eq:transported_graph_context} is computed. The position context in Eq.~\eqref{eq:position_context} is computed independently from graph-predicted node positions and does not come from the transport plan. The two contexts are concatenated in Eq.~\eqref{eq:correction_features}. The gate in Eq.~\eqref{eq:bounded_correction} is one scalar per spatial token and is broadcast over its feature dimensions.

\subsection{Global-Vector Alignment}

For a global visual state, Eqs.~\eqref{eq:global_attention} and \eqref{eq:global_alignment} define attention pooling and the alignment loss. The global visual-state change is the query, graph-node changes are the keys, and the resulting weights pool the unprojected node changes. Since the projected changes are $\ell_2$-normalized, the half-squared $\ell_2$ alignment is equivalent to cosine distance. This adapter does not perform singleton Sinkhorn.

The global correction receives the original global prediction, its predicted change, the pooled graph-node change, and the node features and predicted positions concatenated in fixed task-role order. It applies the same bounded residual principle as the spatial-token path.

\section{Detailed Training Objective}
\label{sec:supp_training_objective}

The complete training objective is
\begin{equation}
\begin{aligned}
\mathcal{L}
={}&
\mathcal{L}_{\mathrm{WM}}
+\lambda_{\mathrm{obs}}\mathcal{L}_{\mathrm{obs}}
+\lambda_{g,1}\mathcal{L}_{g,1}
+\lambda_{\mathrm{align}}\mathcal{L}_{\mathrm{align}}\\
&+\lambda_{\mathrm{var}}\mathcal{L}_{\mathrm{var}}
+\lambda_{\mathrm{emb},K}\mathcal{L}_{\mathrm{emb},K}
+\lambda_{\mathrm{state},K}\mathcal{L}_{\mathrm{state},K}\\
&+\lambda_{\mathrm{pos}}\mathcal{L}_{\mathrm{pos}}
+\lambda_{c,1}\mathcal{L}_{c,1}
+\lambda_{c,K_c}\mathcal{L}_{c,K_c}.
\end{aligned}
\label{eq:total_objective}
\end{equation}
The terms supervise the original visual model, scene-graph observation and prediction, graph--visual alignment, node positions, and visual correction; each $\lambda$ is a non-negative weight. For a global latent state, $\mathcal{L}_{\mathrm{global}}$ replaces $\mathcal{L}_{\mathrm{align}}$. The stop-gradient operator $\operatorname{sg}(\cdot)$ identifies a fixed target that is not updated by the corresponding loss.

\subsection{Visual and Graph Targets}

For DINO-WM, the optional motion weight for visual token $p$ is
\begin{equation}
\begin{aligned}
\widetilde{w}_{t,p}
&=
\left\|
\operatorname{sg}\!\left(
\mathbf{z}_{t+1,p}-\mathbf{z}_{t,p}
\right)
\right\|_2+\varepsilon_m,\\
w_{t,p}
&=
\frac{
\widetilde{w}_{t,p}
}{
P^{-1}\sum_{q=1}^{P}\widetilde{w}_{t,q}
}.
\end{aligned}
\label{eq:supp_motion_weight}
\end{equation}
Here, $\varepsilon_m>0$ retains a contribution from static tokens, and $q$ indexes the tokens in the denominator. The DINO-WM token loss applies weighted Smooth-$L_1$ error between original prediction $\overline{\mathbf{z}}_{t+1,p}$ and fixed target $\mathbf{z}_{t+1,p}$. DreamerV3 retains its native recurrent state-space-model objective. We denote the original visual model objective by $\mathcal{L}_{\mathrm{WM}}$.

Let $b_{t,i}\in\{0,1\}$ indicate whether role $i$ is present and supervised. The observer loss $\mathcal{L}_{\mathrm{obs}}$ uses masked Smooth-$L_1$ losses for physical-state readouts and image positions, binary cross-entropy for presence, and cross-entropy for relation classes. Simulator annotations supervise the observer but are not its inputs.

The $K$-step graph-embedding loss is
\begin{equation}
\mathcal{L}_{\mathrm{emb},K}
=
\frac{
\displaystyle
\sum_{k=1}^{K}
\sum_{i=1}^{J}
b_{t+k,i}
\left\|
\widehat{\mathbf{n}}_{t+k,i}
-
\operatorname{sg}(\mathbf{n}_{t+k,i})
\right\|_2^2
}{
\displaystyle
d_g
\sum_{k=1}^{K}
\sum_{i=1}^{J}
b_{t+k,i}
+\varepsilon_b
},
\label{eq:supp_multistep_graph_loss}
\end{equation}
where $\varepsilon_b>0$ prevents division by zero. The one-step graph loss $\mathcal{L}_{g,1}$ is its $K=1$ counterpart. When physical-state annotations are available, $\mathcal{L}_{\mathrm{state},K}$ supervises an auxiliary readout from predicted nodes. The position loss $\mathcal{L}_{\mathrm{pos}}$ supervises observed, one-step predicted, and recursively predicted node positions.

\subsection{Recursive Correction Training}

The one-step correction loss $\mathcal{L}_{c,1}$ compares the corrected prediction with a fixed next-state target. For spatial visual tokens, the $K_c$-step correction loss is
\begin{equation}
\begin{aligned}
\mathcal{L}_{c,K_c}
={}&
\frac{1}{K_cP}
\sum_{k=1}^{K_c}
\sum_{p=1}^{P}
w_{t+k-1,p}\\[-2pt]
&\quad\cdot
\operatorname{SL1}\!\left(
\widehat{\mathbf{z}}_{t+k,p},
\operatorname{sg}(\mathbf{z}_{t+k,p})
\right).
\end{aligned}
\label{eq:supp_correction_rollout_loss}
\end{equation}
The corrected prediction is fed into the following visual transition. The original visual world model and graph dynamics are evaluated without gradients during this correction rollout, so the loss updates only the correction pathway.

The $K_v=8$ control follows a matched recursive schedule but disables graph correction and feeds the visual model its own predictions. It tests whether longer visual training alone explains JointGV's improvement.

\section{Implementation and Evaluation Details}
\label{sec:supp_evaluation}

\subsection{Implementation Settings}

The primary DMC runs were trained on an NVIDIA GeForce RTX 5090 D v2 GPU with 24\,GB of memory; additional seeds were trained on RTX 4090 and 22\,GB RTX 2080 Ti GPUs using the same model and optimization settings. All DMC tasks use eight graph slots of dimension $d_g=64$, three graph message-passing layers, and a graph-history window of $H_g=4$. Graph dynamics and correction use $K=K_c=8$. Node-feature increments and visual corrections are bounded by $c_n=0.25$ and $c_{\max}=1.0$.

For spatial-token alignment, $d_c=32$, the entropic regularization coefficient is $\epsilon=0.1$, Sinkhorn runs for 20 iterations, and $\sigma_{\mathrm{pos}}=0.15$. The alignment-loss weight starts at 0.1 and is adjusted every 50 minibatches using an exponential moving average of gradient norms. DINO-WM and DreamerV3 use a learning rate of $10^{-4}$ and batch size 16. DINO-WM uses AdamW with 16-frame sequences; DreamerV3 uses Adam with 64-frame sequences. DINOv2-S/14 and RelTR inputs are resized to $224\times224$ pixels. 

We first verify the DreamerV3 implementation using its online actor--critic protocol; the retained Walker Walk agent achieves an evaluation return of \(737.8\pm15.2\). We then freeze this policy and use it to collect a fixed dataset for the prediction experiments. In each data split, 80\% of the episodes are collected with this policy and 20\% use uniformly random actions to broaden the motion coverage. The DreamerV3 backbone and JointGV correction reported in this paper are trained offline from scratch on these fixed trajectories; the online checkpoint is used only for data collection. This protocol ensures that the original and corrected predictions are trained and evaluated on identical observations and actions, without confounding the comparison with online exploration, actor learning, or changes in the replay distribution.

\subsection{Rollout Protocol}

All compared methods receive the same observed frames and recorded actions. A rollout begins with three preceding RGB frames and the rollout-origin frame, followed by 32 recursive predictions. DINO-WM uses three visual-history frames and JointGV uses four graph-history frames. DreamerV3 filters the four observed context frames through its recurrent state-space-model posterior. No future RGB observation, graph, or simulator state is supplied after prediction begins.

The Cheetah Run and Walker Walk test sets each contain 20 episodes and 14 fixed windows per episode, giving 280 matched windows per task. Stored RGB images are $64\times64$ pixels. Predictions within an experiment use the same decoder for visualization and image-based evaluation.

\subsection{Metric Computation}

At horizon $H$, each metric is averaged over prediction steps 1 through $H$ within a window and then across windows. RGB MSE compares a decoded prediction with its target image. SSIM uses an $11\times11$ Gaussian window with standard deviation 1.5. LPIPS uses its AlexNet feature network. DINOv2-token MSE compares predicted tokens with frozen-encoder tokens extracted from the observed future image.

Joint-position error measures articulated structure. The same frozen RGB observer extracts normalized two-dimensional positions from every predicted RGB frame; distances to target positions are averaged over visible nodes and converted to pixels. This evaluation observer was trained separately from JointGV's internal graph observer and does not access JointGV's graph features, predicted positions, or relations. Applying it to observed DMC frames gives measurement floors of 0.19 pixels for Cheetah Run and 1.46 pixels for Walker Walk. Graph-only errors use graph-predicted positions directly and therefore have a different measurement path.

\subsection{Computational Cost}
\label{sec:supp_compute}

The DINO-WM backbone contains 22.7M parameters, of which 0.326M are trainable. JointGV adds 64.4M parameters through the graph branch and its RelTR-based observer, of which 0.531M are trainable. JointGV also increases rollout time because the graph observer, recursive graph dynamics, and correction are evaluated alongside the visual backbone. The graph observer is evaluated once on the context frames, while recursive graph dynamics and correction account for most of the additional cost.

\section{Additional Experiments and Ablation Studies}
\label{sec:supp_ablation}

\subsection{Visual Training, Graph Prediction, and Gradient Coupling}

Increasing the visual training horizon from $K_v=1$ to $K_v=8$ improves the original DINO-WM prediction at later steps, but JointGV further improves either model (Table~\ref{tab:dinowm_results}).

We also test whether the correction loss should update the original visual dynamics. The detached plug-in uses $\omega=0$, while partial and full coupling use $\omega=0.25$ and $\omega=1$. Table~\ref{tab:supp_coupling} separately reports the original visual output and corrected output from each coupled run. Gradient coupling does not improve upon the detached plug-in and degrades the original visual output. Graph-only rows measure recursive graph prediction before graph information is transferred to the visual state.

\begin{table*}[t]
\centering
\footnotesize
\setlength{\tabcolsep}{4pt}
\begin{tabular}{@{}lccccccc@{}}
\toprule
& \multicolumn{3}{c}{Joint-position error (px) $\downarrow$}
& DINOv2 token MSE $\downarrow$
& RGB MSE$\times10^{3}$ $\downarrow$
& SSIM $\uparrow$
& LPIPS $\downarrow$ \\
\cmidrule(lr){2-4}
\cmidrule(lr){5-5}
\cmidrule(lr){6-6}
\cmidrule(lr){7-7}
\cmidrule(lr){8-8}
Output
& $H{=}8$ & $H{=}16$ & $H{=}32$
& $H{=}32$ & $H{=}32$ & $H{=}32$ & $H{=}32$ \\
\midrule
Detached plug-in: corrected output ($\omega{=}0$)
& 0.730 & 0.796 & 0.852 & 0.341 & 1.84 & 0.861 & 0.0351 \\
Full gradient coupling: original visual output ($\omega{=}1$)
& 0.992 & 1.184 & 1.340 & 0.903 & 2.96 & 0.786 & 0.0592 \\
Full gradient coupling: corrected output ($\omega{=}1$)
& 0.738 & 0.812 & 0.882 & 0.381 & 1.97 & 0.853 & 0.0374 \\
Partial gradient coupling: original visual output ($\omega{=}0.25$)
& 0.925 & 1.039 & 1.153 & 0.785 & 2.58 & 0.813 & 0.0532 \\
Partial gradient coupling: corrected output ($\omega{=}0.25$)
& 0.771 & 0.838 & 0.890 & 0.449 & 1.99 & 0.846 & 0.0364 \\
\midrule
Graph-only prediction ($K_v{=}1$)
& 0.520 & 0.609 & 0.704 & -- & -- & -- & -- \\
Graph-only prediction ($K_v{=}8$)
& 0.535 & 0.630 & 0.754 & -- & -- & -- & -- \\
\bottomrule
\end{tabular}
\caption{\textbf{Gradient-coupling and graph-only controls.} ``Original visual output'' is the native prediction from the visual model trained with the indicated correction-gradient fraction; ``corrected output'' includes JointGV. Graph-only errors are measured directly from graph-predicted positions and are not ranked against RGB-observer errors.}
\label{tab:supp_coupling}
\end{table*}

\subsection{Correction and Change Alignment}

The visual-only controls test whether residual capacity or additional node supervision explains the improvement. Mean graph-motion broadcasting tests whether the predicted graph rollout is useful without token-specific coupling. The cosine and OT variants then test two operators for coupling graph-node and visual-token changes; Table~\ref{tab:supp_paired_ablation} reports the matched comparisons.

\begin{table}[t]
\centering
\scriptsize
\setlength{\tabcolsep}{2.6pt}
\renewcommand{\arraystretch}{0.96}
\begin{tabular}{@{}lccc@{}}
\toprule
Correction variant
& Own backbone
& Corrected
& Reduction \\
\midrule

\multicolumn{4}{@{}l}{\emph{Residual controls}} \\
Visual-only residual
& 0.8978 & 0.8935 & 0.0043 \\
Visual-only residual + node supervision
& 0.8978 & 0.8830 & 0.0148 \\

\addlinespace[1pt]
\multicolumn{4}{@{}l}{\emph{Graph-conditioned correction}} \\
Mean graph-motion broadcast
& 0.8978 & 0.8146 & 0.0832 \\
Cosine change coupling
& 0.9021 & 0.7934 & 0.1087 \\
OT change coupling
& 0.8978 & 0.7906 & 0.1072 \\

\bottomrule
\end{tabular}
\caption{\textbf{Matched within-run correction ablations on DMC Cheetah Run.}
Values are joint-position errors averaged over rollout steps 1--32. Each corrected result is compared with the visual backbone from the same training run; Reduction denotes the absolute decrease from that backbone. The visual-only controls test residual capacity and matched node supervision without providing a predicted graph. Mean broadcasting tests the graph correction signal without token-specific coupling, while cosine and OT couple graph-node and visual-token changes.}
\label{tab:supp_paired_ablation}
\end{table}

Across three seeds, the visual-only controls provide substantially smaller improvements than JointGV. Mean graph-motion broadcasting also improves the backbone, showing that the predicted graph rollout contains useful correction information. Cosine and OT coupling provide a further reduction by assigning differentiated graph motion to the visual tokens, while their similar accuracy shows that this interface does not depend on a single coupling operator.

\subsection{Inter-Node Message Passing}

The no-message-passing control disables observer IMP and cross-node message passing in graph dynamics. Its corrected 32-step joint-position error is 0.8588 pixels, compared with 0.9646 pixels for its own backbone. Full JointGV reaches 0.7906 pixels with its own backbone at 0.8978 pixels. The no-message-passing run therefore has higher absolute error, although its separately trained backbone prevents the difference from isolating only node-to-node communication.

% \subsection{Global Node Context}

% For DreamerV3, the ablation removes the fixed-order global context containing graph-node features and predicted positions while retaining attention-pooled graph motion and global change alignment. Table~\ref{tab:dreamerv3_results} reports its own DreamerV3 backbone. The correction improves this backbone at every horizon. Because the full and no-context variants are separate runs, their corrected outputs alone do not isolate the effect of the additional global node context.

\subsection{Training Seeds and Paired Bootstrap}
\label{sec:supp_multiseed}

For DINO-WM on Cheetah Run, we repeat JointGV and the visual-only residual control with three independently trained seeds. Each corrected prediction is compared with the backbone from the same run on the same 280 test windows. We use 2,000 paired bootstrap resamples of these windows within each seed. These intervals measure test-window sampling variation, whereas the standard deviation across seeds measures training variation.

\begin{table}[t]
\centering
\scriptsize
\setlength{\tabcolsep}{2.5pt}
\renewcommand{\arraystretch}{0.95}
\resizebox{\columnwidth}{!}{
\begin{tabular}{llcccc}
\toprule
Method & Seed & Backbone & Corrected & Reduction & 95\% CI \\
\midrule
JointGV & 0 & 0.8978 & 0.7906 & 11.95\% & [11.08, 12.72]\% \\
JointGV & 1 & 0.9377 & 0.7790 & 16.92\% & [16.17, 17.70]\% \\
JointGV & 2 & 0.9899 & 0.8001 & 19.18\% & [18.20, 20.10]\% \\
JointGV & Mean$\pm$std
& 0.9418$\pm$0.0462
& 0.7899$\pm$0.0106
& 16.01$\pm$3.70\%
& -- \\
\midrule
Visual-only residual & 0 & 0.8978 & 0.8935 & 0.49\% & [-0.27, 1.23]\% \\
Visual-only residual & 1 & 0.9516 & 0.9194 & 3.38\% & [2.46, 4.35]\% \\
Visual-only residual & 2 & 0.9998 & 0.9094 & 9.04\% & [8.45, 9.63]\% \\
Visual-only residual & Mean$\pm$std
& 0.9497$\pm$0.0510
& 0.9074$\pm$0.0131
& 4.30$\pm$4.35\%
& -- \\
\bottomrule
\end{tabular}
}
\caption{\textbf{Three-seed results on the 280 fixed DMC Cheetah Run test windows.} Backbone and corrected columns report average joint-position error over rollout steps 1--32. Reduction is computed relative to the corresponding backbone from the same run. Confidence intervals are obtained by paired bootstrap resampling within each seed; mean and standard deviation summarize variation across independently trained seeds.}
\label{tab:supp_multiseed}
\end{table}

Table~\ref{tab:supp_multiseed} shows that JointGV improves its corresponding backbone in every seed and produces a larger and more consistent reduction than the visual-only residual. The visual-only residual also improves two of the three runs, but its reduction is smaller and more variable; its seed-0 confidence interval includes zero. We also repeat the primary DreamerV3--Walker Walk comparison with three seeds. JointGV reduces the average 32-step joint-position error from $5.835\pm0.093$ to $4.336\pm0.019$ pixels, corresponding to a relative reduction of $25.7\pm1.3\%$. The reductions for seeds 0, 1, and 2 are $27.0\%$, $25.7\%$, and $24.4\%$, respectively.

\subsection{Reversed Backbone--Task Pairings}

We additionally evaluate DINO-WM on Walker Walk and DreamerV3 on Cheetah Run, reversing the backbone--task pairings in the main comparisons. Table~\ref{tab:supp_cross_task} reports representative matched runs for all four pairings.

\begin{table*}[t]
\centering
\footnotesize
\setlength{\tabcolsep}{3.8pt}
\begin{tabular}{@{}llcccccc@{}}
\toprule
Model / task & Output
& \multicolumn{3}{c}{Joint-position error (px) $\downarrow$}
& RGB MSE$\times10^{3}$ $\downarrow$
& SSIM $\uparrow$
& LPIPS $\downarrow$ \\
\cmidrule(lr){3-5}
\cmidrule(lr){6-6}
\cmidrule(lr){7-7}
\cmidrule(lr){8-8}
& & $H{=}8$ & $H{=}16$ & $H{=}32$
& $H{=}32$ & $H{=}32$ & $H{=}32$ \\
\midrule
DINO-WM / Cheetah Run
& Original
& 0.830 & 0.931 & 1.036 & 2.31 & 0.841 & 0.0435 \\
& + JointGV
& 0.730 & 0.796 & 0.852 & 1.84 & 0.861 & 0.0351 \\
\midrule
DreamerV3 / Cheetah Run
& Original
& 0.378 & 0.427 & 0.470 & 0.978 & 0.929 & 0.0178 \\
& + JointGV
& 0.355 & 0.394 & 0.425 & 0.798 & 0.939 & 0.0153 \\
\midrule
DINO-WM / Walker Walk
& Original
& 7.807 & 9.626 & 10.953 & 13.52 & 0.568 & 0.2705 \\
& + JointGV
& 6.368 & 6.996 & 7.671 & 11.74 & 0.625 & 0.2382 \\
\midrule
DreamerV3 / Walker Walk
& Original
& 4.063 & 4.750 & 5.938 & 15.12 & 0.621 & 0.1722 \\
& + JointGV
& 3.785 & 4.022 & 4.338 & 11.15 & 0.694 & 0.1231 \\
\bottomrule
\end{tabular}
\caption{\textbf{Representative matched runs for all four DMC model--task pairings.} Values at horizon $H$ average rollout steps 1 through $H$. The three-seed joint-position results are summarized in Table~\ref{tab:dreamerv3_results}.}
\label{tab:supp_cross_task}
\end{table*}

Across three independently trained seeds, JointGV reduces the average 32-step joint-position error of DINO-WM on Walker Walk from $10.740\pm0.922$ to $7.682\pm0.151$ pixels, corresponding to a relative reduction of $28.0\pm7.8\%$. For DreamerV3 on Cheetah Run, the error decreases from $0.467\pm0.009$ to $0.415\pm0.009$ pixels, corresponding to a reduction of $11.1\pm2.6\%$. The improvement occurs in every seed for both reversed pairings.

Representative rollouts for DreamerV3 on Cheetah Run and DINO-WM on Walker Walk are shown in Figs.~\ref{fig:supp_dreamerv3_cheetah} and~\ref{fig:supp_dinowm_walker}, respectively.

\subsection{Per-Step and Per-Window Analyses}

\begin{figure}[t]
\centering
\includegraphics[width=\linewidth]{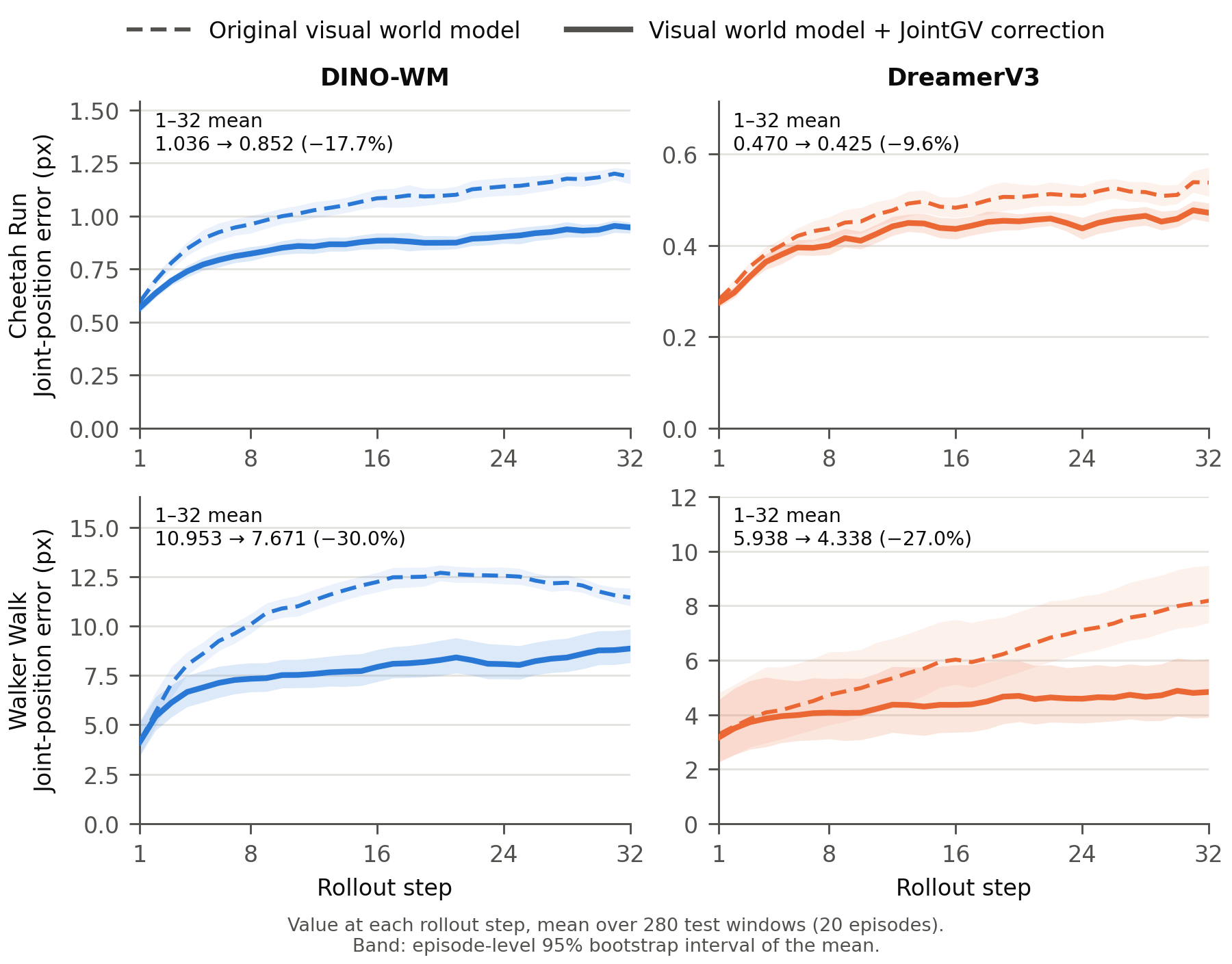}
\caption{Per-step joint-position error for one run of each DINO-WM and DreamerV3 task pairing. Dashed curves show the original predictions and solid curves show JointGV-corrected predictions. Shaded regions are episode-level 95\% bootstrap intervals over the fixed test windows. JointGV reduces structural error over most of the rollout rather than only at its final step.}
\label{fig:supp_per_step_joint}
\end{figure}

The per-step curves in Fig.~\ref{fig:supp_per_step_joint} show that JointGV's improvement is distributed across the rollout rather than arising only at the final prediction step. The corresponding per-window distributions are reported in Fig.~\ref{fig:supp_paired_windows} of the main paper.

% \begin{figure}[t]
% \centering
% \includegraphics[width=\linewidth]{figures/supp_per_window_violin.png}
% \caption{\textbf{Distribution of per-window changes for one run of each pairing.}Each value is the relative change from the original prediction to JointGV, with joint-position error averaged over rollout steps 1--32; values below zero indicate improvement. The bars show the interquartile ranges, the white dots show the medians, and the percentages report the shares of test windows improved by JointGV.}
% \label{fig:supp_paired_windows}
% \end{figure}

\subsection{Supervision Scope}

JointGV uses simulator-provided node identities, visibility, positions, and task relations as training targets for graph construction and graph prediction. The visual-only residual in Table~\ref{tab:dinowm_results} receives neither graph features nor node annotations. It therefore tests whether a generic bounded residual explains the gain, but does not by itself isolate the effect of additional structured supervision.

To control for supervision volume, we additionally train a visual-only residual with an auxiliary node head supervised by the same MuJoCo node annotations as JointGV. The correction module does not receive the predicted node states or graph features. The supervision-matched residual reduces the average 32-step joint-position error from $0.8978$ to $0.8830$ pixels, a $1.65\%$ reduction with a paired-bootstrap 95\% interval of $[0.95,2.33]\%$. JointGV reduces the same backbone error to $0.7906$ pixels, corresponding to an $11.95\%$ reduction. The additional node supervision alone therefore does not explain JointGV's gain.

\subsection{Stratified Analysis}
\label{sec:supp_stratified}

We analyze when graph--visual coupling provides the largest additional benefit. For each DINO-WM test window, we compute two quantities over rollout steps 1--32: (i) the minimum ground-truth image distance between non-adjacent nodes, which measures how closely separate articulated parts approach in the image, and (ii) the mean error of the graph-predicted node positions. The 280 windows of each task are divided into four equal-sized groups according to each quantity.

Table~\ref{tab:supp_stratified} reports paired differences in average 32-step joint-position error. For Cheetah Run, we compare mean graph-motion broadcasting with OT change coupling to isolate the additional benefit of token-specific graph--visual coupling. For Walker Walk, we compare JointGV with its original visual backbone to examine how structural ambiguity and graph-prediction quality affect the complete correction. Confidence intervals use 2,000 paired bootstrap resamples over windows. Because windows from the same episode are not independent, these window-level intervals are optimistic.

\begin{table*}[t]
\centering
\footnotesize
\setlength{\tabcolsep}{4pt}
\begin{tabular}{@{}lcccc@{}}
\toprule
Paired difference (px)
& Q1
& Q2
& Q3
& Q4 \\
\midrule

\multicolumn{5}{@{}l}{
\emph{Minimum distance between non-adjacent nodes (Q1: closest)}
} \\
Cheetah Run: mean broadcast $-$ OT
& $+0.038$ $[+0.024,+0.051]$
& $+0.036$ $[+0.025,+0.046]$
& $+0.022$ $[+0.010,+0.035]$
& $+0.001$ $[-0.013,+0.014]$ \\
Walker Walk: JointGV $-$ original
& $-3.92$ $[-4.28,-3.52]$
& $-3.39$ $[-3.80,-2.97]$
& $-3.34$ $[-3.80,-2.84]$
& $-2.48$ $[-3.10,-1.88]$ \\

\midrule
\multicolumn{5}{@{}l}{
\emph{Graph-predicted position error (Q1: most accurate)}
} \\
Cheetah Run: mean broadcast $-$ OT
& $+0.025$ $[+0.012,+0.038]$
& $+0.024$ $[+0.011,+0.037]$
& $+0.030$ $[+0.017,+0.043]$
& $+0.017$ $[+0.004,+0.030]$ \\
Walker Walk: JointGV $-$ original
& $-4.69$ $[-4.92,-4.46]$
& $-4.34$ $[-4.58,-4.10]$
& $-3.07$ $[-3.52,-2.59]$
& $-1.02$ $[-1.50,-0.57]$ \\

\bottomrule
\end{tabular}
\caption{\textbf{Stratified paired differences in average 32-step joint-position error.}
Each task contains 280 DINO-WM seed-0 test windows, with 70 windows per quartile. Brackets give paired-bootstrap 95\% confidence intervals. Positive mean-broadcast $-$ OT values indicate lower error with OT coupling; negative JointGV $-$ original values indicate lower error after correction.}
\label{tab:supp_stratified}
\end{table*}

On Cheetah Run, the additional gain from OT coupling over mean broadcasting is concentrated in windows where non-adjacent nodes approach in image space. The difference decreases from $0.038$ pixels in the closest quarter to $0.001$ pixels in the farthest quarter, with Spearman correlation $\rho=-0.22$ and a 95\% interval of $[-0.33,-0.10]$. Its relation to graph-position error is weak ($\rho=-0.06$, $[-0.17,0.06]$), indicating that this coupling advantage is associated more strongly with projected structural ambiguity than with overall node-position accuracy.

On Walker Walk, JointGV also provides its largest reduction when non-adjacent nodes are closest. Its improvement decreases as graph-predicted position error grows: the paired difference changes from $-4.69$ pixels in the most accurate quarter to $-1.02$ pixels in the least accurate quarter. The correlation between graph-position error and the JointGV-minus-original difference is $\rho=0.67$, with a 95\% interval of $[0.59,0.73]$. This result identifies inaccurate graph prediction, particularly under severe limb overlap, as a remaining failure mode.

\subsection{Observer Robustness and Independent Evaluation}
\label{sec:supp_observer}

The internal graph observer is used only to initialize the graph history from the observed RGB context. To measure sensitivity to initialization errors, we add Gaussian noise \(\mathcal{N}(0,\sigma^2)\) to the normalized RGB seed frames read by the internal observer, using \(\sigma\in\{0,0.02,0.05,0.1,0.2,0.3\}\). The visual backbone, target frames, and evaluation observer always receive clean images. After initialization, the graph is predicted recursively and receives no additional image observations.

\begin{figure}[t]
\centering
\includegraphics[width=\columnwidth]{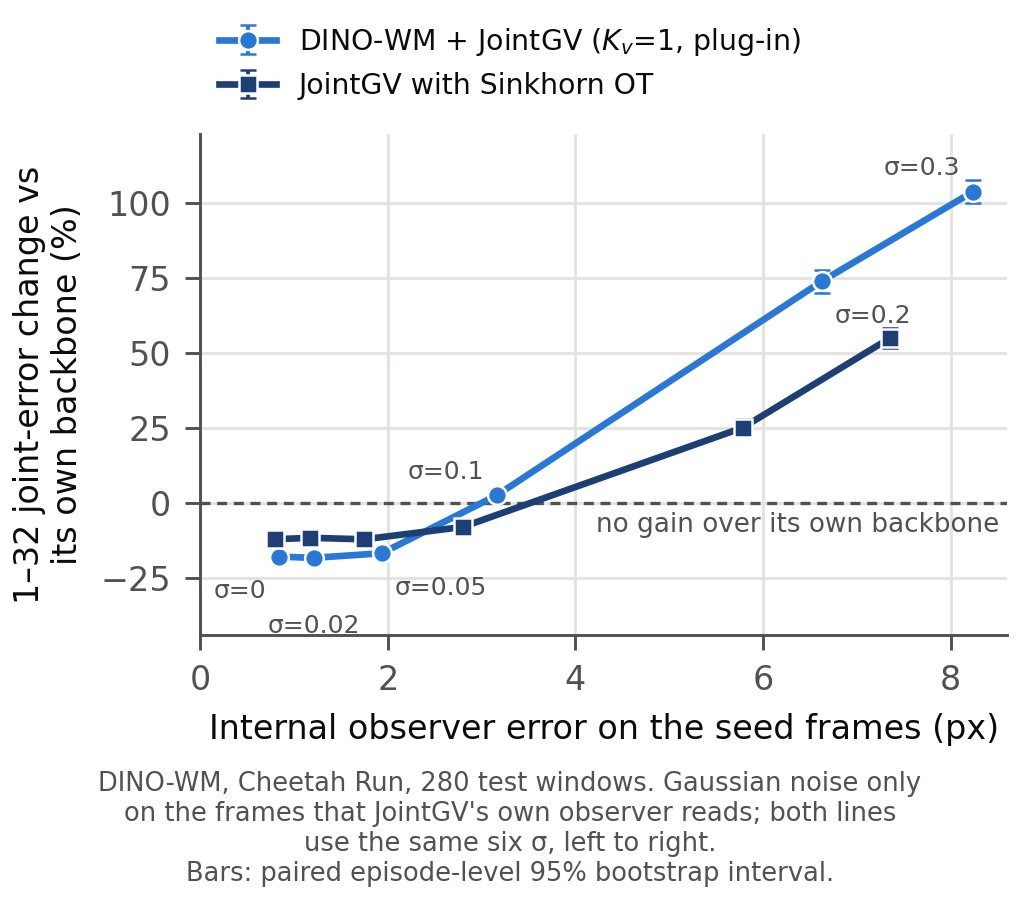}
\caption{\textbf{Sensitivity to errors in JointGV's internal graph observer.} Gaussian noise is applied only to the RGB context read by this observer; the visual backbone and evaluation observer receive clean images. Negative values indicate lower error than the corresponding backbone. Bars show paired episode-level 95\% bootstrap intervals.}
\label{fig:supp_observer_noise}
\end{figure}

Figure~\ref{fig:supp_observer_noise} shows that both configurations remain more accurate than their corresponding backbones through $\sigma=0.05$. At $\sigma=0.10$, JointGV with Sinkhorn OT retains a $7.9\%$ reduction in 32-step joint-position error, whereas the primary plug-in becomes $2.7\%$ worse than its backbone. Both configurations degrade under the larger perturbations. JointGV therefore tolerates moderate graph-initialization error but currently lacks a fallback when the observed graph becomes severely inaccurate.

The main joint-position metric is produced by a frozen RelTR-based RGB observer trained separately from JointGV. Because this evaluator and the internal graph observer originate from the same pretrained architecture, we additionally rescore the same predicted RGB rollouts using an independently trained heatmap-based keypoint network. This network is trained from scratch on observed training images and does not use RelTR features or model-generated images.

On observed target frames, the independent evaluator has a measurement error of $0.093$ pixels, compared with $0.196$ pixels for the RelTR-based evaluator. It measures an average 32-step joint-position error of $0.786$ pixels for the original DINO-WM prediction and $0.679$ pixels for JointGV, corresponding to a $13.6\%$ reduction with a paired-bootstrap 95\% interval of $[12.7,14.5]\%$; JointGV improves $95.4\%$ of the test windows. Under the same evaluator, the visual-only residual is $3.4\%$ worse than its backbone. The JointGV improvement and the relative ordering of the controls therefore do not depend on using a RelTR-based evaluator.

\subsection{Contact-Rich Manipulation on DMC Finger Turn Hard}

To extend the DMC evaluation beyond articulated locomotion, we evaluate JointGV on Finger Turn Hard, where a two-link finger must contact and rotate a spinner toward a target. The scene graph contains six task roles: the proximal finger, distal finger, fingertip, spinner center, spinner tip, and target. The task introduces robot--object contact, object rotation, and a separate goal entity that are absent from Cheetah Run and Walker Walk.

We evaluate DINO-WM, DreamerV3, and LeWorldModel over three independently trained seeds using the same 280 fixed test windows and 32-step open-loop protocol. Within each seed, the original and corrected predictions use the same visual backbone, selected checkpoint, decoder, observed context, and action sequence; they differ only in whether JointGV correction is enabled. Joint-position error is measured from decoded predictions by the frozen RGB observer over all visible task-role nodes. An independently trained keypoint evaluator, which uses neither RelTR features nor generated images during training, provides an additional measurement for DINO-WM and DreamerV3.

Table~\ref{tab:finger_results} reports the compact DINO-WM comparison in the main paper. Table~\ref{tab:supp_finger_results} gives the complete seed-0 visual metrics, while Table~\ref{tab:supp_finger_multiseed} summarizes the three-seed results for all three visual world models.

\begin{table*}[t]
  \centering
  \footnotesize
  \setlength{\tabcolsep}{3.5pt}
  \renewcommand{\arraystretch}{0.95}
  \begin{tabular}{@{}lccccccc@{}}
    \toprule
     & \multicolumn{3}{c}{Joint-position error (px) $\downarrow$}
     & Token MSE $\downarrow$
     & RGB MSE$\times10^{3}$ $\downarrow$
     & SSIM $\uparrow$
     & LPIPS $\downarrow$ \\
    \cmidrule(lr){2-4}
    \cmidrule(lr){5-5}
    \cmidrule(lr){6-6}
    \cmidrule(lr){7-7}
    \cmidrule(lr){8-8}
    Method
     & $H{=}8$ & $H{=}16$ & $H{=}32$
     & $H{=}32$ & $H{=}32$ & $H{=}32$ & $H{=}32$ \\
    \midrule
    DINO-WM backbone~\cite{zhou2024dinowm}
     & 10.813 & 12.476 & 14.861
     & 2.204 & 11.29 & 0.633 & 0.3673 \\
    DINO-WM + JointGV
     & \textbf{9.444} & \textbf{9.911} & \textbf{10.754}
     & \textbf{0.823} & \textbf{5.44}
     & \textbf{0.761} & \textbf{0.2444} \\
    \bottomrule
  \end{tabular}
  \caption{\textbf{DINO-WM seed-0 results on 280 DMC Finger Turn Hard test windows.} Values at horizon $H$ average rollout steps 1 through $H$. Joint-position error is measured from decoded predictions over visible task-role nodes, and Token MSE denotes DINOv2-token MSE. Both rows use the same checkpoint and differ only in whether JointGV correction is enabled.}
  \label{tab:supp_finger_results}
\end{table*}

\begin{table*}[t]
\centering
\scriptsize
\setlength{\tabcolsep}{2.5pt}
\renewcommand{\arraystretch}{0.95}
\begin{tabular}{@{}lcccccccc@{}}
\toprule
& \multicolumn{3}{c}{Joint error}
& Independent joint
& RGB MSE
& SSIM
& LPIPS
& Motion MSE \\
\cmidrule(lr){2-4}
\cmidrule(lr){5-5}
\cmidrule(lr){6-6}
\cmidrule(lr){7-7}
\cmidrule(lr){8-8}
\cmidrule(lr){9-9}
Visual world model
& $H{=}8$ & $H{=}16$ & $H{=}32$
& $H{=}32$ & $H{=}32$ & $H{=}32$ & $H{=}32$ & $H{=}32$ \\
\midrule
DINO-WM
& $-14.3{\pm}1.5$
& $-22.6{\pm}1.9$
& $-27.5{\pm}1.4$
& $-44.6{\pm}0.7$
& $-53.2{\pm}1.4$
& $+21.4{\pm}1.2$
& $-40.6{\pm}6.2$
& $-18.8{\pm}3.8$ \\
DreamerV3
& $-8.2{\pm}0.4$
& $-11.1{\pm}0.5$
& $-12.0{\pm}2.3$
& $-21.6{\pm}7.1$
& $+1.0{\pm}2.7$
& $+0.0{\pm}0.1$
& $+2.5{\pm}1.9$
& $-3.5{\pm}0.8$ \\
LeWorldModel~\cite{maes2026leworldmodel}
& $-0.5{\pm}0.1$
& $-1.5{\pm}1.1$
& $-3.2{\pm}1.4$
& --
& $-5.9{\pm}1.5$
& $+0.4{\pm}0.2$
& $-2.5{\pm}0.9$
& $-2.8{\pm}1.4$ \\
\bottomrule
\end{tabular}
\caption{\textbf{Three-seed relative changes produced by JointGV on Finger Turn Hard (\%).} Each entry is the mean and standard deviation of the within-seed change from the corresponding visual backbone. Negative values indicate lower joint, RGB, LPIPS, or motion-weighted error; positive SSIM values indicate improvement. Motion MSE weights RGB error by ground-truth image motion. The independent keypoint result is not reported for the LeWorldModel follow-up.}
\label{tab:supp_finger_multiseed}
\end{table*}

The DINO-WM improvement is consistent across seeds and becomes larger with rollout horizon: the average joint-position error reduction increases from $14.3\%$ at $H=8$ to $27.5\%$ at $H=32$. The independent evaluator measures a larger $44.6\pm0.7\%$ reduction at $H=32$, while RGB MSE decreases by $53.2\pm1.4\%$. DreamerV3 also shows consistent structural improvement, reducing the primary and independent joint errors by $12.0\pm2.3\%$ and $21.6\pm7.1\%$, respectively; its aggregate RGB appearance metrics do not change consistently. LeWorldModel has the smallest correction, but its joint-position and RGB errors still decrease across all three seeds. These results extend JointGV from articulated locomotion to contact-rich manipulation and show increasing structural benefit over longer rollouts for three different visual states.

\subsection{Planning with the Official DINO-WM Push-T Checkpoint}

We attach JointGV to the authors' released DINO-WM Push-T checkpoint~\cite{zhou2024dinowm}, leaving the visual model and official CEM/MPC settings unchanged. This avoids confounding the comparison with backbone reproduction or insufficient visual-model training. Only the graph and correction modules are trained over three seeds. This post-hoc attachment differs from the full training schedule used in our preceding experiments and was not its best-performing correction setting, providing a conservative test on a released planning model.

DINO-WM uses an action skip of five; hence $gH=5$ spans 25 environment steps and matches its original planning horizon. We additionally test $gH=10$ and $gH=25$, corresponding to 50 and 125 steps. Each setting uses the same 50 goals, initial states, and planning parameters. Table~\ref{tab:supp_pusht_planning} reports success and final-state distance for each horizon. Final-state distance combines the image-space pusher and T-block target errors and therefore does not measure T-block placement alone.

\begin{table*}[t]
\centering
\footnotesize
\setlength{\tabcolsep}{5pt}
\renewcommand{\arraystretch}{0.95}
\begin{tabular}{@{}lcccc@{}}
\toprule
& \multicolumn{2}{c}{Success (\%) $\uparrow$}
& \multicolumn{2}{c}{Final-state distance (px) $\downarrow$} \\
\cmidrule(lr){2-3}
\cmidrule(lr){4-5}
Goal horizon
& Official
& JointGV s0 / s1 / s2
& Official
& JointGV s0 / s1 / s2 \\
\midrule
$gH{=}5$  & 90 & 90 / 90 / 88 & 13.64  & 12.05 / 11.20 / 11.82 \\
$gH{=}10$ & 10 & 18 / 16 / 16 & 58.58  & 52.56 / 57.97 / 58.69 \\
$gH{=}25$ & 0  & 0 / 0 / 0   & 122.06 & 104.65 / 102.14 / 105.89 \\
\bottomrule
\end{tabular}
\caption{\textbf{Push-T planning using the official DINO-WM checkpoint and three post-hoc JointGV seeds.} Each horizon contains the same 50 goals. The official checkpoint and baseline outputs are identical across seeds; only the JointGV seed changes.}
\label{tab:supp_pusht_planning}
\end{table*}

At the protocol-matched $gH=5$, JointGV preserves the official model's near-saturated success. At $gH=10$, success increases from $5/50$ to $9/50$, $8/50$, and $8/50$, a seed-averaged increase of $6.7$ percentage points with a 95\% confidence interval of $[1.3,13.3]$. Each seed retains all five official successes and solves three or four additional goals.

At $gH=25$, the official DINO-WM latent no longer supports successful planning, so we additionally compare final-state distance. JointGV reduces it from $122.1$ pixels to $104.7$, $102.1$, and $105.9$ pixels, with a 95\% interval of $[-30.9,-4.8]$ pixels for the difference. Most of this reduction comes from the pusher ending closer to its target; the T-block position and orientation do not improve significantly. Thus, the clearest planning gain occurs at $gH=10$, while $gH=25$ shows improved approach rather than completed placement.

\begin{figure*}[t]
\centering
\includegraphics[width=0.7\textwidth]{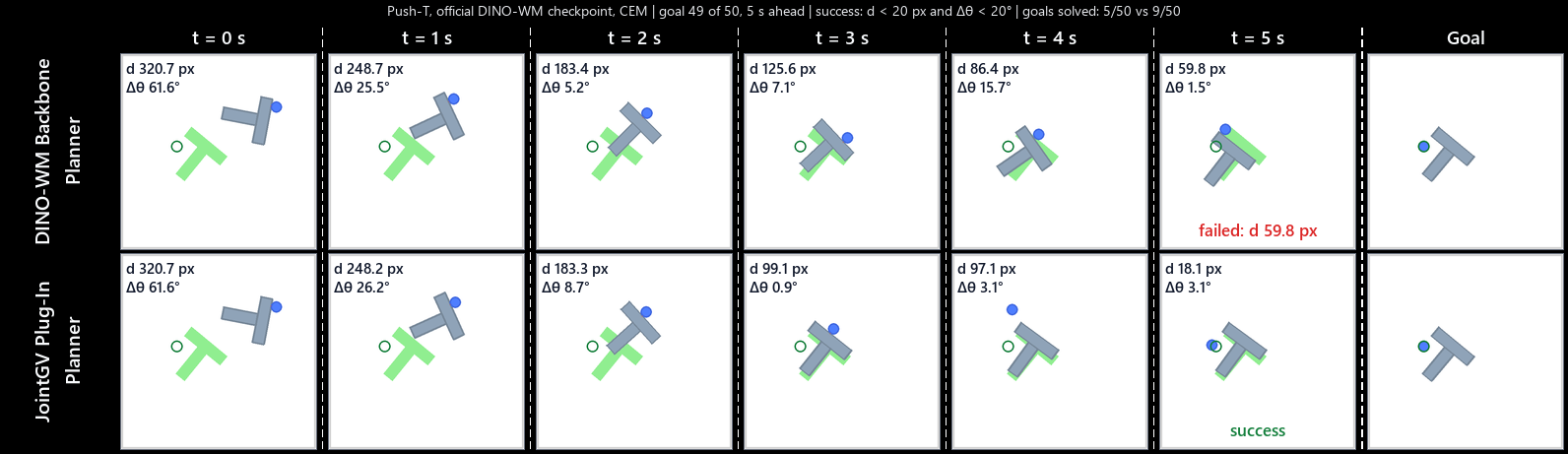}
\caption{\textbf{Push-T planning with the official DINO-WM checkpoint at $gH{=}10$.} The two CEM planners start from the same state and target and differ only in whether their visual rollouts are corrected by JointGV. In this episode, the original planner stops outside the positional success threshold, whereas the JointGV planner reaches the target pose. Across the same 50 goals, the two planners solve 5 and 9 goals, respectively.}
\label{fig:supp_pusht_episode}
\vspace{-5pt}
\end{figure*}

Figure~\ref{fig:supp_pusht_episode} shows a representative $gH{=}10$ episode. Both planners initially move the object toward the goal, but the original planner remains outside the positional threshold, whereas JointGV continues toward the target and satisfies both the position and orientation criteria. This example illustrates that the corrected rollout changes the actions selected by CEM rather than only improving the decoded visualization.

\section{Real-Robot Inverted-Pendulum Prediction and Control}
\label{sec:supp_real_robot}

\subsection{Open-Loop Prediction on Real-Robot Recordings}

We evaluate 156 matched windows from 96 recorded RM65-B inverted-pendulum episodes. For the physical dataset, task-role node targets are generated offline by posing a calibrated RM65-B geometry model from synchronized robot joint angles and SA100 pole-angle measurements, then projecting its representative robot-link and pole points into the image. Color-based pole-tip detections are used only to fit the camera extrinsics and timing and angle offsets. These targets supervise training only; recursive inference uses the observed RGB context and recorded actions. All models receive the same observed context and recorded actions, then predict 32 future steps without future images. Table~\ref{tab:supp_real_cartpole_prediction} reports each visual world model before and after correction.

\begin{table}[H]
\centering
\scriptsize
\setlength{\tabcolsep}{2.5pt}
\renewcommand{\arraystretch}{0.95}
\begin{tabular}{@{}llccc@{}}
\toprule
Model & Output
& Joint (px) $\downarrow$
& Pole (deg) $\downarrow$
& RGB MSE$\times10^3$ $\downarrow$ \\
\midrule
DINO-WM
& Original & 2.590 & 12.564 & 10.659 \\
& + JointGV & 2.124 & 6.689 & 8.067 \\
\addlinespace[1pt]
DreamerV3
& Original & 1.137 & 3.805 & 4.342 \\
& + JointGV & 0.894 & 2.825 & 2.217 \\
\addlinespace[1pt]
LeWorldModel~\cite{maes2026leworldmodel}
& Original & 1.576 & 4.823 & 5.805 \\
& + JointGV & 1.442 & 4.242 & 5.055 \\
\bottomrule
\end{tabular}
\caption{\textbf{Open-loop prediction on 156 matched real-robot inverted-pendulum windows.} Values average rollout steps 1--32; comparisons are within each visual world model. DINO-WM and LeWorldModel use matched checkpoints, while DreamerV3 uses separately selected original and corrected checkpoints.}
\label{tab:supp_real_cartpole_prediction}
\vspace{-4pt}
\end{table}

JointGV reduces all three reported errors for DINO-WM and DreamerV3. LeWorldModel also improves under the same evaluation windows, although its 32-step joint-position and pole-angle reductions are smaller. These are open-loop prediction results, not physical balancing results. Representative real-robot rollouts for the three visual world models are shown in Figs.~\ref{fig:supp_dinowm_real}--\ref{fig:supp_leworldmodel_real}.

% \begin{figure}[t]
% \centering
% \includegraphics[width=\linewidth]{figures/cartpole_actor_seed42003_t0-1-2-3-4-5s.png}
% \caption{\textbf{MuJoCo inverted-pendulum control from one fixed test seed.} The original-model actor reaches the fall threshold after approximately 1.44 seconds; the JointGV actor completes the six-second simulation horizon. Aggregate results are given in Table~\ref{tab:supp_cartpole_control}.}
% \label{fig:supp_robot_simulation}
% \end{figure}

\subsection{MuJoCo Inverted-Pendulum Control and Hardware Status}
\label{sec:supp_robot_simulation}

\begin{figure*}[t]
\centering
\includegraphics[width=0.82\textwidth]{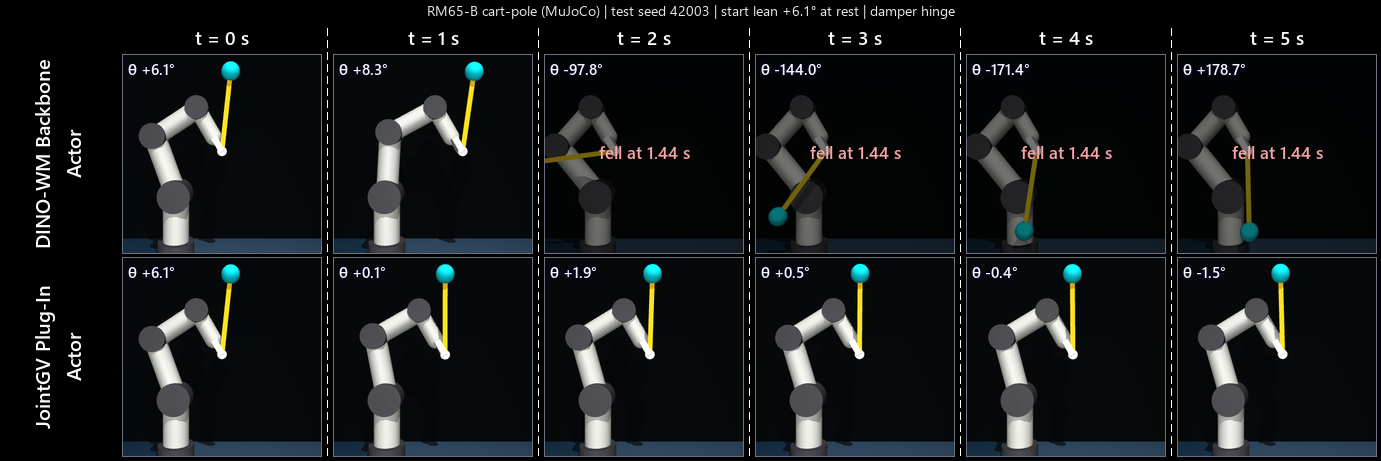}
\caption{\textbf{Representative MuJoCo inverted-pendulum control rollout.}
Both actors start from the same unseen simulator state. The actor trained with original DINO-WM rollouts reaches the fall threshold after approximately $1.44$ seconds, whereas the actor trained with JointGV-corrected rollouts completes the six-second simulation horizon. Aggregate results over 50 unseen initial states are reported in Table~\ref{tab:control} of the main paper.}
\label{fig:supp_robot_simulation}
\vspace{-6pt}
\end{figure*}

We test whether the prediction improvement can support control in a MuJoCo model of the RM65-B inverted-pendulum system. Both actors use the same action constraints and a shared reward predictor trained on observed simulator RGB frames. The task reward decreases with absolute pole angle from vertical, and an episode terminates when the pole reaches the horizontal. The actors differ in the visual predictions used during imagined training: one uses the original DINO-WM prediction, while the other uses JointGV-corrected predictions.

Figure~\ref{fig:supp_robot_simulation} shows one representative rollout, and Table~\ref{tab:control} of the main paper reports the aggregate results. Across 50 unseen simulator initial states, the JointGV actor completes the six-second horizon in 46 trials, compared with none for the original-model actor. Its mean time before termination increases from 2.28 to 5.86 seconds. The shared reward predictor also has lower validation error on JointGV-corrected rollouts, although this is only a diagnostic; balancing is measured by executing each actor in the simulator.

For DINO-WM, JointGV also reduces the average 32-step pole-angle-change error from $3.088^\circ$ to $1.457^\circ$ and obtains the lower error on $82.1\%$ of the matched windows, indicating more accurate prediction of the pole's tilt trend.

We also tested the actors on the physical RM65-B system. The earlier seven paired trials used preliminary checkpoints; both actors reached the angle-triggered stop, although JointGV produced a longer recorded control window in six pairs. The representative trial shown in Fig.~\ref{fig:robot_control} starts from comparable pole angles: after $1.27$ seconds, the original-rollout actor reaches $-83.3^\circ$, whereas the JointGV-rollout actor remains at $-2.7^\circ$. This demonstrates a short-duration physical-control advantage, but not stable balancing. High-speed operation introduces non-negligible command delay, and we have not yet obtained a controller that consistently maintains the physical pole for six seconds. Table~\ref{tab:supp_cartpole_control} summarizes these hardware trials.

\begin{table}[H]
\centering
\footnotesize
\setlength{\tabcolsep}{3pt}
\renewcommand{\arraystretch}{0.95}
\begin{tabular}{@{}lcc@{}}
\toprule
Hardware: seven paired trials & Original & + JointGV \\
\midrule
Angle-triggered stops & 7/7 & 7/7 \\
Mean recorded control window (s) & 0.85 & 0.98 \\
\bottomrule
\end{tabular}
\caption{\textbf{Preliminary hardware control trials.} Hardware trials use earlier actor checkpoints and are not directly comparable with the MuJoCo results in Table~\ref{tab:control}.}
\label{tab:supp_cartpole_control}
\vspace{-6pt}
\end{table}

\section{Additional Qualitative Rollouts}
\label{sec:supp_qualitative}

\begin{figure*}[!p]
\centering
\includegraphics[width=0.8\textwidth]{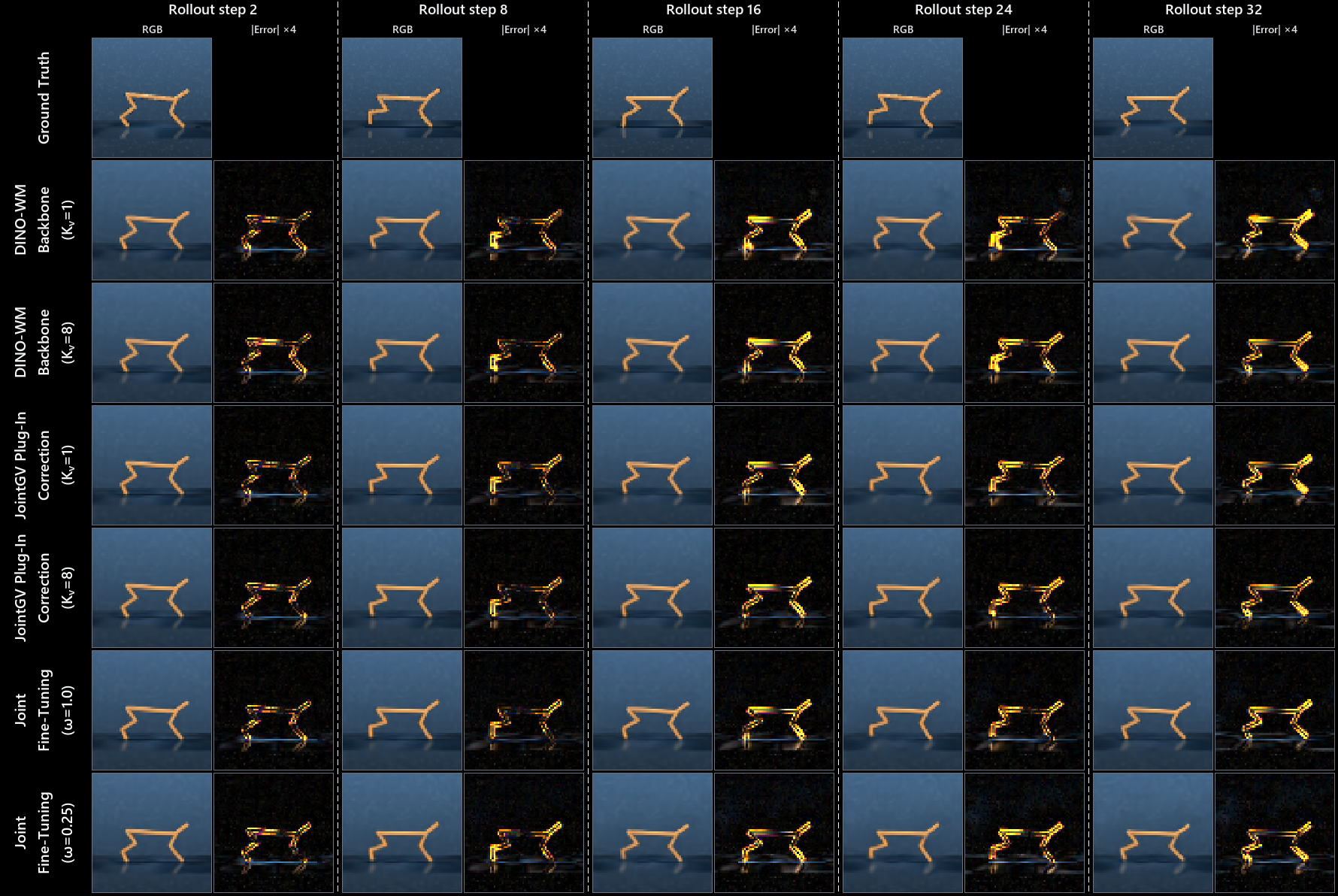}
\caption{\textbf{Additional DINO-WM rollouts on three held-out DMC Cheetah Run test episodes.} Rows, displayed steps, decoder, and error-map scaling follow Fig.~\ref{fig:dinowm_qualitative}. JointGV generally preserves articulated shape more consistently than the corresponding original prediction. Occasional local failures are retained.}
\label{fig:supp_dinowm_rollouts}
\vspace{-5pt}
\end{figure*}

\begin{figure*}[!p]
\centering
\includegraphics[width=0.8\textwidth]{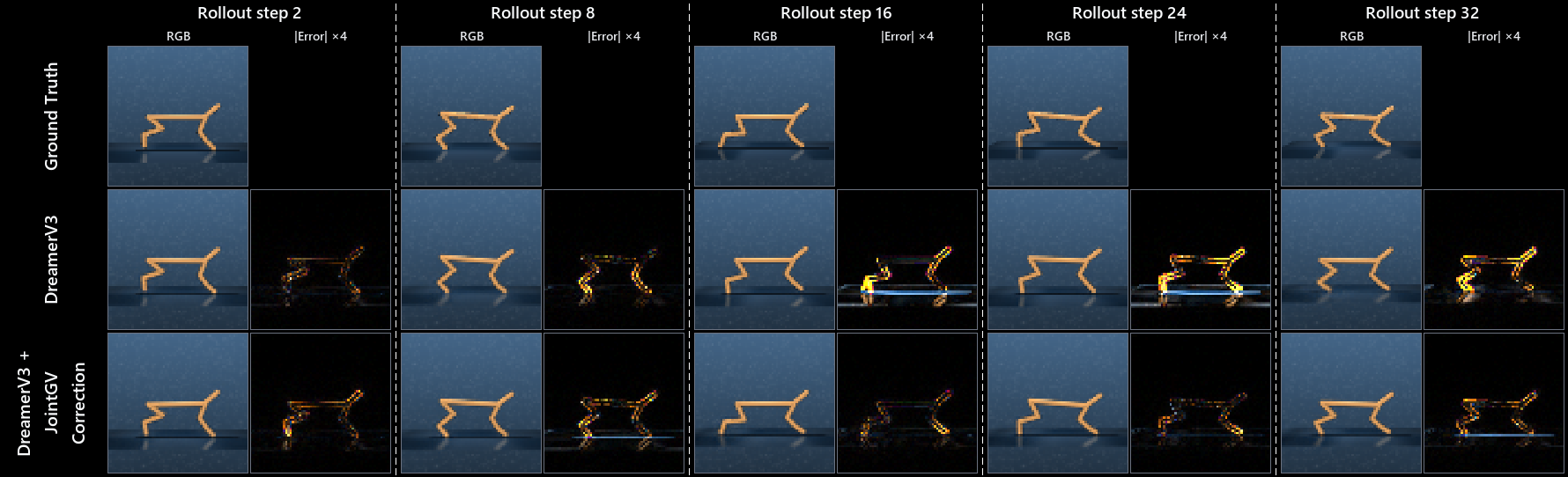}
\caption{\textbf{DreamerV3 rollout on a held-out DMC Cheetah Run episode.} JointGV better preserves the articulated configuration and reduces visual error as the rollout progresses, showing that the correction also applies to this reversed backbone--task pairing.}
\label{fig:supp_dreamerv3_cheetah}
\vspace{-5pt}
\end{figure*}

\begin{figure*}[!p]
\centering
\includegraphics[width=0.8\textwidth]{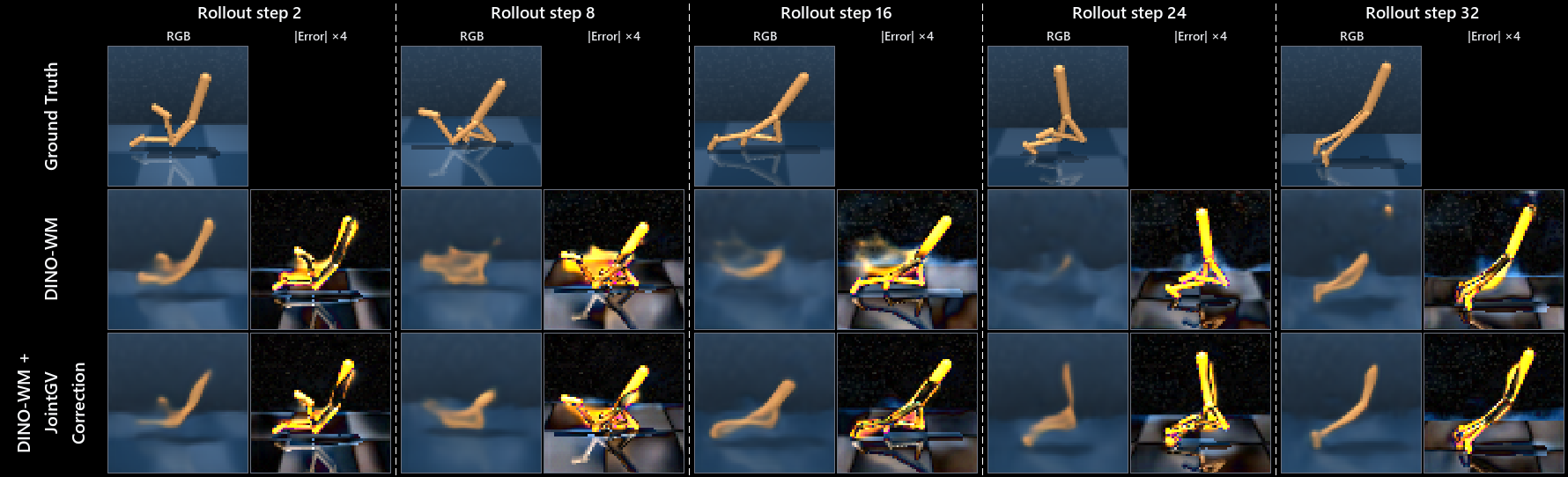}
\caption{\textbf{DINO-WM rollout on a held-out DMC Walker Walk episode.} The original prediction becomes increasingly blurred and loses limb definition, whereas JointGV retains a clearer articulated pose over the rollout.}
\label{fig:supp_dinowm_walker}
\vspace{-5pt}
\end{figure*}

\begin{figure*}[!p]
\centering
\includegraphics[width=0.8\textwidth]{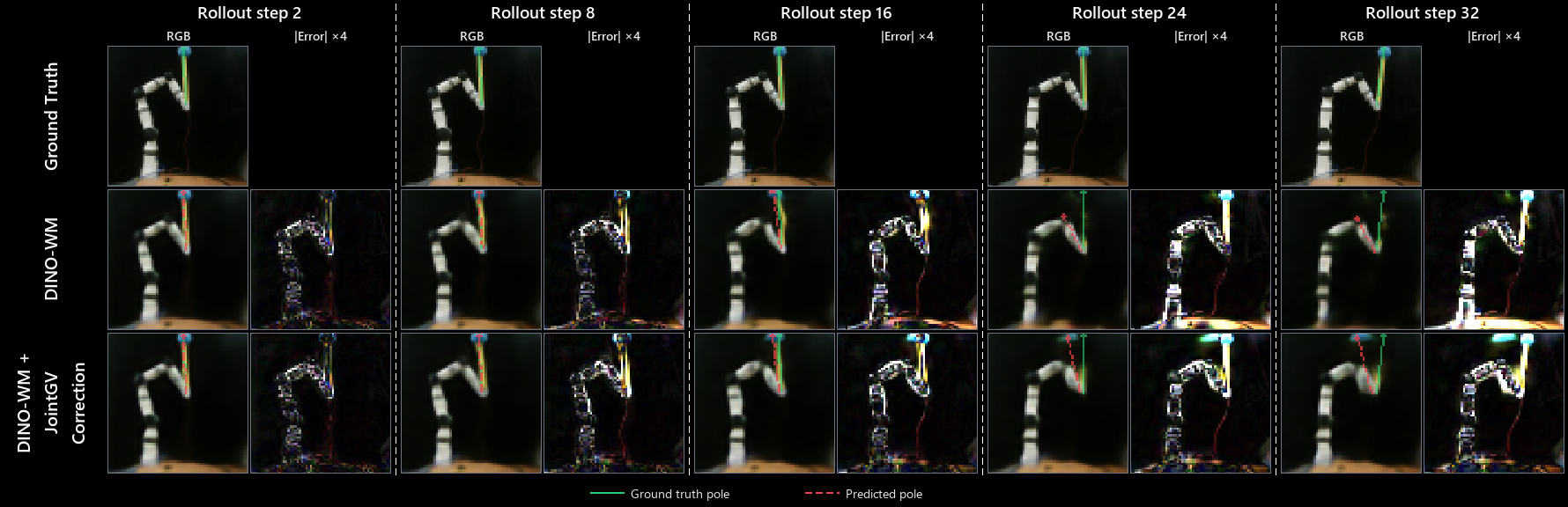}
\caption{\textbf{DINO-WM rollout on a held-out real-robot inverted-pendulum sequence.} The pole-axis overlays show how orientation error develops across the rollout. JointGV improves several intermediate predictions, although residual late-horizon deviation remains in this selected sequence.}
\label{fig:supp_dinowm_real}
\vspace{-5pt}
\end{figure*}

\begin{figure*}[!p]
\centering
\includegraphics[width=0.8\textwidth]{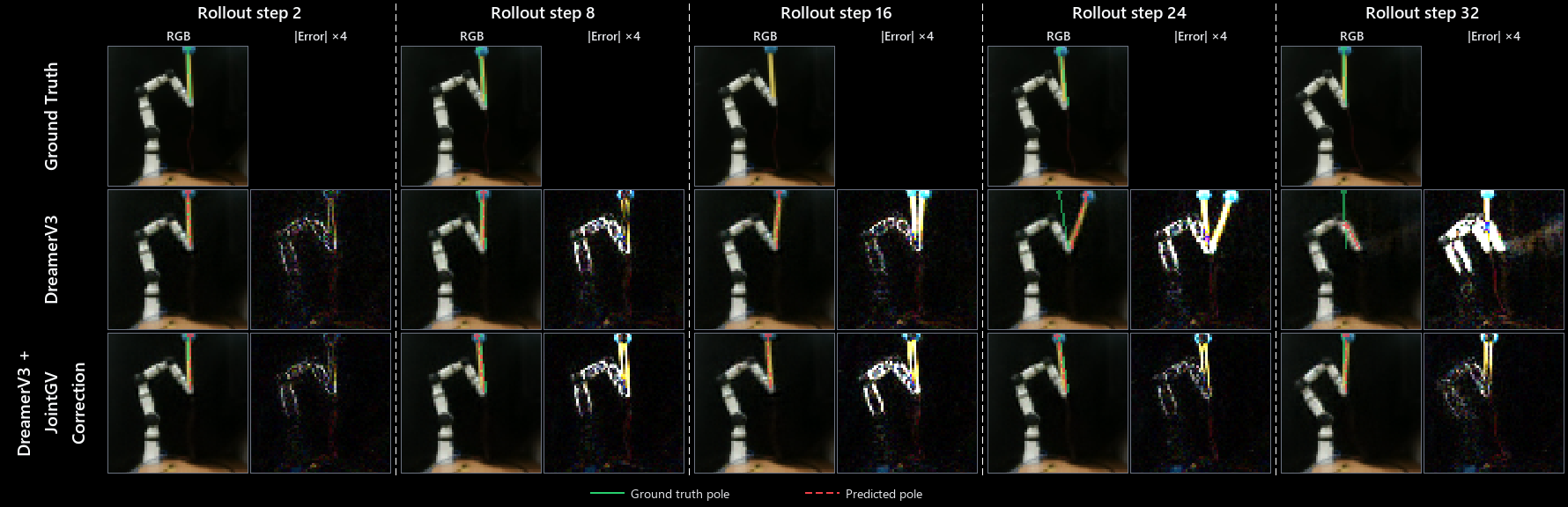}
\caption{\textbf{DreamerV3 rollout on a held-out real-robot inverted-pendulum sequence.} JointGV more closely follows the observed pole orientation at later rollout steps, consistent with the aggregate reduction in pole-angle error.}
\label{fig:supp_dreamerv3_real}
\vspace{-5pt}
\end{figure*}

\begin{figure*}[!p]
\centering
\includegraphics[width=0.8\textwidth]{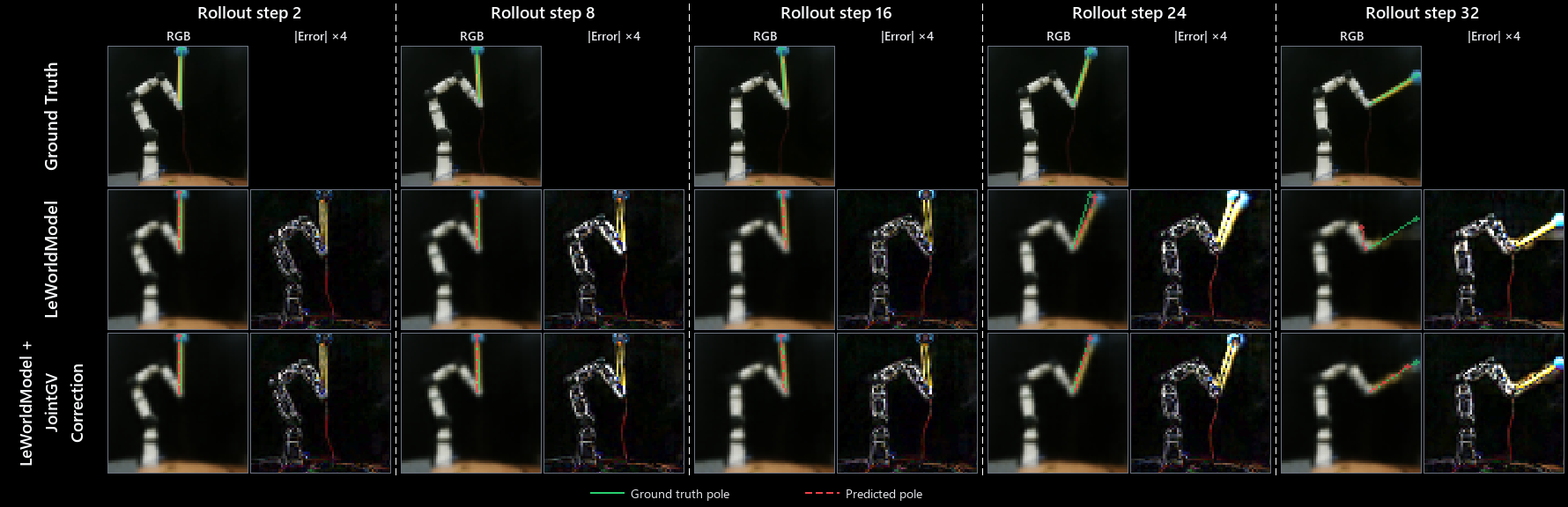}
\caption{\textbf{LeWorldModel rollout on a held-out real-robot inverted-pendulum sequence.} The original model already retains much of the pole structure, and JointGV provides a smaller correction than for DINO-WM and DreamerV3, consistent with the aggregate results in Table~\ref{tab:supp_real_cartpole_prediction}.}
\label{fig:supp_leworldmodel_real}
\vspace{-5pt}
\end{figure*}

Figure~\ref{fig:supp_dinowm_rollouts} shows three additional DINO-WM rollouts on held-out Cheetah Run episodes, including occasional local failures. Figures~\ref{fig:supp_dreamerv3_cheetah} and~\ref{fig:supp_dinowm_walker} provide qualitative results for the reversed DMC backbone--task pairings.  Figures~\ref{fig:supp_dinowm_real}--\ref{fig:supp_leworldmodel_real} show representative predictions on real-robot inverted-pendulum recordings.

\paragraph{Reversed DMC backbone--task pairings.}
The following panels show predictions at rollout steps 2, 8, 16, 24, and 32. Each predicted RGB frame is followed by its absolute RGB error magnified by four. These examples complement the quantitative reversed-pairing results in Sec.~\ref{sec:supp_ablation}.

\paragraph{Real-robot inverted-pendulum rollouts.}
All three models receive the same observed RGB context and recorded actions and predict 32 future steps without access to future images. The solid green and dashed red lines indicate the observed and predicted pole orientations, respectively. These selected sequences complement the aggregate results in Table~\ref{tab:supp_real_cartpole_prediction}.

% Compact spacing for the final panel-only pages
\makeatletter
\setlength{\@dblfptop}{0pt}
\setlength{\@dblfpsep}{6pt}
\setlength{\@dblfpbot}{0pt plus 1fil}
\makeatother
\setlength{\abovecaptionskip}{3pt}
\setlength{\belowcaptionskip}{0pt}
\renewcommand{\dblfloatpagefraction}{0.5}

\begin{figure*}[!p]
\centering
\includegraphics[width=0.8\textwidth]{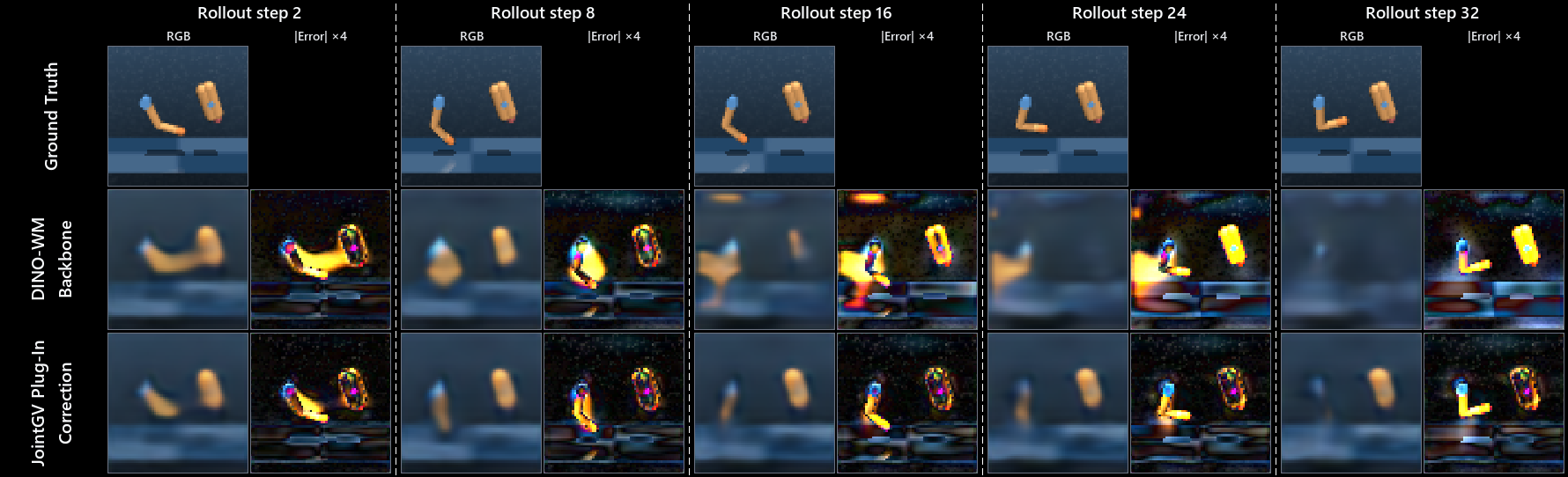}
\caption{\textbf{DINO-WM rollout on a held-out DMC Finger Turn Hard episode.} The original prediction progressively loses the finger and spinner geometry, whereas JointGV better preserves their structure during the contact-rich rollout.}
\label{fig:supp_finger_rollout}
\vspace{-5pt}
\end{figure*}

\paragraph{Supplementary video.}
The accompanying video presents complete 32-step rollouts for DINO-WM and DreamerV3 on Cheetah Run and Walker Walk, including three test episodes for each model--task pair. It additionally shows two matched real-robot prediction episodes for each of DINO-WM, DreamerV3, and LeWorldModel, followed by a MuJoCo inverted-pendulum control demonstration and two physical-robot control episodes comparing the original and JointGV-corrected actors. Each prediction rollout displays ground-truth RGB, the original visual world model, and its JointGV-corrected counterpart using the same decoder and error-map scaling as the corresponding static figures.